\documentclass[acmsmall]{acmart}
\AtBeginDocument{%
  \providecommand\BibTeX{{%
    \normalfont B\kern-0.5em{\scshape i\kern-0.25em b}\kern-0.8em\TeX}}}

\setcopyright{acmlicensed}
\acmJournal{TELO}
\acmYear{2025} \acmVolume{1} \acmNumber{1} \acmArticle{1} \acmMonth{1}\acmDOI{10.1145/3715012}

\acmBooktitle{Woodstock '18: ACM Symposium on Neural Gaze Detection,
 June 03--05, 2018, Woodstock, NY} 
\acmISBN{978-1-4503-XXXX-X/18/06}

\usepackage{siunitx}
\usepackage{algorithm, algorithmic}
\usepackage{subcaption}
\usepackage{mathptmx}

\theoremstyle{plain}
\newtheorem{theorem}{Theorem}[section]

\theoremstyle{definition}
\newtheorem{definition}[theorem]{Definition}

\theoremstyle{remark}

\newcommand{\R}{\mathbb{R}}
\newcommand{\E}{\mathbb{E}}

\begin{document}

\title{Combining Multi-Objective Bayesian Optimization with Reinforcement Learning for TinyML}

\author{Mark Deutel}
\email{mark.deutel@fau.de}
\affiliation{%
  \institution{Friedrich-Alexander-Universität Erlangen-Nürnberg}
  \postcode{91058}
  \city{Erlangen}
  \country{Germany}
}

\author{Georgios Kontes}
\email{georgios.kontes@iis.fraunhofer.de}
\affiliation{%
  \institution{Fraunhofer IIS, Fraunhofer Institute for Integrated Circuits IIS}
  \postcode{91058}
  \city{Erlangen}
  \country{Germany}
}

\author{Christopher Mutschler}
\email{christopher.mutschler@iis.fraunhofer.de}
\affiliation{%
  \institution{Fraunhofer IIS, Fraunhofer Institute for Integrated Circuits IIS}
  \postcode{91058}
  \city{Erlangen}
  \country{Germany}
}

\author{Jürgen Teich}
\email{juergen.teich@fau.de}
\affiliation{%
  \institution{Friedrich-Alexander-Universität Erlangen-Nürnberg}
  \postcode{91058}
  \city{Erlangen}
  \country{Germany}
}


\begin{abstract}
Deploying deep neural networks (DNNs) on microcontrollers (TinyML) is a common trend to process the increasing amount of sensor data generated at the edge, but in practice, resource and latency constraints make it difficult to find optimal DNN candidates. Neural architecture search (NAS) is an excellent approach to automate this search and can easily be combined with DNN compression techniques commonly used in TinyML. However, many NAS techniques are not only computationally expensive, especially hyperparameter optimization (HPO), but also often focus on optimizing only a single objective, e.g., maximizing accuracy, without considering additional objectives such as memory requirements or computational complexity of a DNN, which are key to making deployment at the edge feasible.
In this paper, we propose a novel NAS strategy for TinyML based on multi-objective Bayesian optimization (MOBOpt) and an ensemble of competing parametric policies trained using Augmented Random Search (ARS) reinforcement learning (RL) agents.
Our methodology aims at efficiently finding tradeoffs between a DNN's predictive accuracy, memory requirements on a given target system, and computational complexity. Our experiments show that we consistently outperform existing MOBOpt approaches on different datasets and architectures such as ResNet-18 and MobileNetV3.
\end{abstract}





\maketitle

\section{Introduction}
\label{section:introduction}

The deployment of DNNs on embedded devices is restricted to the constraints imposed by the target microcontroller platform. Constraints such as $<$1MB Flash, $<$512Kb RAM, and clock speeds in the low-MHz range make the design of DNN models for such platforms challenging. In general, several conflicting objectives and constraints such as memory availability, inference time, and power of the deployed DNN model must be considered.

Despite extensive research on neural architecture search (NAS), there is still no definitive method that is both fast and robust. Many black-box optimization approaches~\citep{zoph_neural_2017,real2019regularized,white2021bananas} work reliably, but are inefficient. Differentiable NAS strategies such as DARTS~\citep{liu2018darts} have robustness problems, as it can be seen that the architectures found often do not generalize well because the search overfits the validation dataset~\citep{zela2020understanding}. Finally, zero-cost NAS strategies that have been proposed to allow rapid network specialization for different target platforms and resource constraints, such as Once-for-all~\citep{Cai2020Once-for-All} or PreNAS~\citealp{wang2023prenas}, while fast, do not provide accurate information about an architecture's performance, but only statistics, such as the number of FLOPs~\citep{white2022deeper}.

In this work, we investigate how to improve the efficiency of NAS for DNN deployment on microcontrollers using a combination of multi-objective Bayesian optimization (MOBOpt), deep compression, and Augmented Random Search (ARS)~\citep{mania2018ars}. To this end, we formulate the architecture search as a hyperparameter optimization (HPO) problem~\citep{bergstra2013making}, with the hyperparameters controlling the use of filter pruning~\citep{li_pruning_2017} and weight quantization~\citep{gholami2021survey}. As optimization objectives, we consider top-1 accuracy, memory (ROM and RAM), and FLOPS.

\begin{figure}
    \centering
    \includegraphics[width=.95\textwidth]{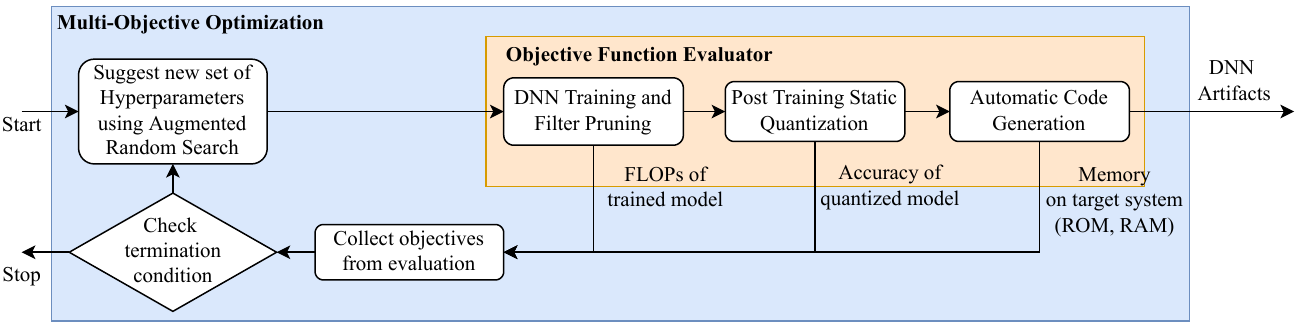}
    \caption{Overview of our proposed RL-based multi-objective Bayesian DNN hyperparameter optimization approach for microcontroller targets.}
    \Description{Our multi-objective DNN hyperparameter optimization approach for microcontrollers uses Augmented Random Search RL agents to iteratively propose new DNN configurations, which in turn are evaluated using objective function evaluators for the four considered objectives: FLOPs, accuracy, ROM, and RAM.}
    \label{fig:moopt_overview}
\end{figure}

Our proposed HPO method consists of two building blocks: (1) the multi-objective optimization loop and (2) the objective functions evaluator, see Fig.~\ref{fig:moopt_overview}. Our proposed search strategy iteratively determines new sets of training, pruning, and quantization hyperparameters until a termination condition is met. The trained DNN resulting from the use of each of the proposed sets of hyperparameters is then evaluated, resulting in a vector of scalar objective values.

Our main contribution, a novel Bayesian-optimization based solver, uses an ensemble of competing locally parameterizable policies that are iteratively trained on the underlying surrogate model using ARS RL agents. Our experiments show that this RL-based strategy helps to improve the effectiveness of our NAS approach in finding optimal DNNs for microcontroller deployment compared to other state-of-the-art multi-objective optimization strategies for multiple use cases while introducing negligible additional overhead for training the RL agents. The architectures proposed by our approach can be deployed directly on common microcontrollers without further (re-)training. To make the results presented in this work accessible, we have open-sourced the implementation of our method\footnote{A Python implementation of the Bayesian optimization-based solver discussed in this work, including the implementation of both the ARS and the PPO RL agents, is available at \url{https://doi.org/10.5281/zenodo.13918411}. The repository includes documentation on how to install and execute the code.}.

The remainder of the paper is structured as follows. Section~\ref{section:related-work} discusses related work. Section~\ref{section:background} provides background on multi-objective Bayesian optimization. Section~\ref{section:method} describes our RL agents and how we integrate them into our MOBOpt DNN hyperparameter exploration process. We evaluate our approach in Section~\ref{section:evaluation}. First, we compare our MOBOpt approach with other optimizers (Bayesian and non-Bayesian) on several public datasets and with two types of DNN, MobileNetV3 and ResNet18, see Section~\ref{sec:algobench}. Second, in Section~\ref{sec:ablation} and~\ref{sec:intuition}, we evaluate the robustness and performance of the ARS agents by discussing the hyperparameter selection and results on a small synthetic optimization problem. Third, we compare ARS with PPO, an alternative deep-RL algorithm, and compare our approach to other optimization strategies such as Automated Pruning and Zero-Shot NAS in Section~\ref{sec:compare}.
\section{Related Work}
\label{section:related-work}

Different approaches for efficient design and deployment of DNNs on embedded platforms have been prominently discussed in scientific research: Dedicated architectures such as the MobileNet class~\citep{howard2017mobilenets, sandler2018mnetv2, howard2019searching} have been proposed, introducing scaling parameters and specialized layers, i.e., depthwise-separable convolutions, to control size and inference time.
Deep compression features a number of techniques, including pruning~\citep{li_pruning_2017}, which dynamically removes weights during training, and weight quantization~\citep{jacob_quantization_2017}, which reduces the resolution at which weights are represented. Knowledge distillation (KD)~\citep{hinton2015distilling} has also been used for DNN compression~\citep{ashok2018n2n,cao2018learnable}. However, DNN optimization using KD remains challenging, as the student often fails to match the predictions of the teacher~\citep{stanton2021does,cho2019efficacy}. Based on DNN compression deployment pipelines for embedded devices have been proposed: \citet{han_deep_2016} combine network pruning, weight quantization, and Huffman coding, and follow-up work compares different pruning and quantization methods~\citep{deutel_deployment_2022}. Other well-known DNN deployment pipelines are MCUNet~\citep{lin_mcunet_2020,lin2021mcunetv2} and Tensorflow Lite Micro~\citep{david2021tensorflow}.

Although NAS is an intensively researched topic, harnessing its power to find DNN deployment candidates for microcontroller platforms is still a largely unsolved problem. Previous approaches~\citep{wang2023prenas,Cai2020Once-for-All,cai2018proxylessnas} focus mainly on zero-cost NAS that derive architecture variants for a given target platform and a given set of resource constraints from a well-trained "supernet" structure or focus on hardware-aware search for larger mobile platforms~\citep{tan2019mnasnet,dong2018dpp}. In contrast, although largely unexplored in the context of NAS for microcontroller targets, multi-objective Bayesian Optimization (MOBOpt) is one of the most widely used black-box optimization methods in NAS to solve HPO problems.~\citep{elsken2018efficient,white2021bananas,chitty2022neural,white2023neural}. As DNN training is time- and resource-intensive the reduced sample complexity of Bayesian optimization is beneficial compared to otherwise popular genetic optimization~\citep{lu2019nsga, deb2002nsgaii}. When optimizing continuous hyperparameters, algorithms using Gaussian processes (GP) are most prominent~\citep{knowles2006parego}. An extension of this optimization approach that specifically aims to tackle problems with high-dimensional search spaces suggests the use of locally modeled confidence regions~\citep{eriksson2019turbo,daulton2022multiobjective}.

The use of RL for MOOpt has been extensively studied: The general approach is to consider the search space as the action space of the RL agent, whose reward is the performance of the trained architecture on unseen data. Approaches to optimization that use RL differ in how they represent the policies they use, e.g., \citet{he2018amc} use Deep Deterministic Policy Gradient (DDPG) agents, \citet{JMLR_v15_vanmoffaert14a} use tabular Q-learning, \citet{KIM2022111263} use single-step RL approaches, and \citet{li2020deep} utilize actor-critic algorithms. In this work, we combine the sampling efficiency of GPs with the expressive power of RL agents. Compared to previous work on RL for MOOpt, our approach does not directly apply RL to the search space of the MOOpt problem, but instead uses RL to improve the optimization performed on the GPs. To address the problem of expending large amounts of resources to train RL agents, instead we propose to train a set of competing policies within MOBOpt that implement candidate sampling using very simple Augmented Random Search~(ARS)~\cite{mania2018ars} RL agents. Hence, our approach allows for efficient exploration of DNN architectures that can be directly deployed on microcontrollers.

Besides MOOpt, RL has been used extensively for automatic DNN pruning as well. Pruning strategies such as AMC~\cite{he2018amc}, DECORE~\cite{alwani2022decore} or NEON~\cite{hirsch2022multi} use RL to suggest sparsity ratios per layer for a base model to perform pruning. Either a single RL agent is applied iteratively layer by layer, suggesting sparsity ratios for each prunable layer individually (AMC and NEON), or a different RL agent is used for each layer, resulting in a multi-agent approach (DECORE). As a result, the RL agents replace the tuning of pruning hyperparameters usually performed by a human engineer. Such automatic pruning approaches differ significantly from our approach and from MOOpt in general in that they do not explore trade-offs between different pruning schedules and hyperparameter configurations while iteratively learning from successive separate DNN training and pruning processes, but instead automatically search for a sensible pruning configuration as part of a single run. As a result, automatic pruning does not produce a Pareto front like MOOpt would, nor does it consider other hyperparameters that influence training, such as training epochs, batch size, or learning rate, but is limited to exploring sparsity ratios alone.

As an alternative to offline pruning, which is performed as part of DNN training, dynamic pruning describes methods that select a subset of layer structures of a DNN at runtime based on the processed inputs~\cite{wang2018skipnet,gao2018dynamic,liu2018dynamic,bolukbasi2017adaptive}. Compared to MOOpt as presented in this work, where a Pareto front of different and unique architectural tradeoffs between accuracy, memory, and computational complexity is searched offline, dynamic pruning never actually removes structures of an architecture completely, but instead dynamically selects different subsets based on the input it receives online. As a result, dynamic pruning only reduces the computational effort of DNN inference, but not the problem of limited memory (RAM, ROM) as considered here in the context of TinyML.
\section{Background}
\label{section:background}

The goal of multi-objective optimization (MOOpt) is to either maximize or minimize a set of objectives. Assume, without loss of generality, maximizing some set of objective functions $f(x) = [f_1(x),\dots,f_n(x)] \in \R^n$, where $n \geq 2$, while satisfying a set set of constraints $g(x) \geq 0 \in \R^V$ where $V \geq 0$, $x \in \mathcal{X} \subset \R^d$, and $\mathcal{X}$ is a compact set. Usually, there exists no single solution $x^*$ that maximizes all objectives while also satisfying all $V$ constraints.

\begin{definition}
An objective function evaluation $f(x)$ \textit{Pareto-dominates} $f(x')$, denoted as $f(x) \succ f(x')$, if $f_m(x) \geq f_m(x')$ for all $m = 1,\dots,M$ and there exists at least one $m \in \{1,\dots,M\}$ such that $f_m(x) > f_m(x')$.~\citep{daulton2022multiobjective}
\end{definition}

\begin{definition}
A set of Pareto-optimal tradeoffs $P(x)$ over a set of samples $X \subseteq \mathcal{X}$ is called the \textit{Pareto front} (PF). $\mathcal{P}(X) = \{f(x): x \in X, \nexists x' \in X$ s.t.\ $f(x') \succ f(x) \}$. The \textit{feasible PF} is defined as $\mathcal{P}_{\text{\textit{feas}}}(X) = \mathcal{P}(\{x \in X: g(x) \geq 0\})$.~\citep{daulton2022multiobjective} 
\end{definition}

Hence, the goal of MOOpt is to identify an approximate feasible PF $\mathcal{P}_{feas}(X_n)$ of the true PF $\mathcal{P}(X)$ within a search budget of $n$ evaluations. In cases where the true PF is not known, the quality of $\mathcal{P}_{feas}(X_n)$ is usually assessed using the Hypervolume (HV) indicator, also called the S-metric.

\begin{definition}
The \textit{Hypervolume indicator} $HV(\mathcal{P}(X)|r)$ is the $M$-dimensional Lebesgue measure $\lambda_M$ of the region dominated by $\mathcal{P}(X)$ and bounded from below by a reference point $r \in \R^M$.~\citep{daulton2022multiobjective} 
\end{definition}

A feasible reference point is commonly derived from the domain knowledge of the problem. The HV also allows one to compare multiple PFs as long as they were calculated with the same reference point and share the same objective values.

A way of improving the sample efficiency of multi-objective optimization is by using Bayesian Optimization (BO) aiming at minimizing the number of required evaluations of the given parameter space. BO treats the objective functions as black-boxes and places a prior over them, thereby capturing beliefs about their behavior. Over time, as new samples are collected, the prior is iteratively updated to form the posterior distribution over the objective functions, in our case using GPs. The posterior, in turn, is used in each iteration to determine the next sample-point to evaluate by employing an acquisition function as a heuristic to quantize the "usefulness" of the sample. Traditionally, the acquisition function is also responsible for balancing the exploration and exploitation tradeoff. Furthermore, since we are considering multiple objectives, we scale and accumulate the objective values to form a single-objective problem~\citep{knowles2006parego,paria2020flexible}.

A common acquisition function is the Expected Improvement~(EI)~\cite{mockus1975ei}, see Eq.~\eqref{eq:ei},
\begin{equation}
    \label{eq:ei}
    EI(x)=\mathbb{E}\max(f(x)-f^*,0)
\end{equation}
where $f^*$ is the objective value of the best observed sample so far. For a given Bayesian model the EI can be evaluated with an integral over the posterior distribution either analytically or by Monte-Carlo sampling \cite{wilson2017reparamforaqfunc}, which allows to approximate the EI as shown in Eq.~\eqref{eq:mcei},
\begin{equation}
    \label{eq:mcei}
    qEI(x)\approx\frac{1}{N}\sum_{i=1}^{N}\max_{j\dots q}\{\max(\xi_{i}-f^*,0)\}, \xi_i\sim\mathbb{P}(f(x)\mid\mathcal{D}),
\end{equation}
where $q$ is the number of samples considered jointly and $\mathbb{P}(f(x)\mid\mathcal{D})$ is the posterior distribution of the function $f$ at $x$ given the data $\mathcal{D}$ observed so far.

\begin{figure}[t]
\begin{algorithm}[H]
\caption{Multi-Objective Bayesian Optimization with Augmented Random Search}
\Description{The Multi-Objective Bayesian Optimization with Augmented Random Search that we propose in this paper and that we describe in detail in the Methods section of this paper.}
\label{alg:nn_ars}
\begin{algorithmic}
   \STATE {\bfseries Parameters:} search budged $J$, learning rate $\alpha$, directions sampled per iteration $N$, rollout horizon $H$, constant exploration noise $v$, number of top-performing directions to use $b$, number of objectives $n$, number of parameters $p$, number of ARS agents $L$, objective function $f(x) = [f_1(x),\dots,f_n(x)]$
   \STATE Create initial prior by evaluating $f(x)$ several times using Latin-Hypercube sampling
   \FOR{$i \gets 1$ {\bfseries to} $J$}
   \STATE Fit GPs by maximising the marginal likelihood of previous evaluations of $f(x)$.
   \STATE Select $\{x_1, x_2, \dots, x_{L}\}$ initial states from previous evaluations $\mathcal{P}_{feas}$ using k-means clustering.
   \REPEAT
   \FOR{$l \gets 0$ {\bfseries to} $L$}
   \STATE Sample directions $\varphi_1, \varphi_2, \dots, \varphi_{N} \in \mathbb{R}^{p \times n}$ with i.i.d. standard normal entries.
   \STATE Collect the summed reward of the MC-sampled and scaled posteriors, see Equations~\ref{eq:scale} and \ref{eq:reward}, of $2N$ rollouts over the horizon $H$ from the GP using local policy $\pi_l$ and initial state $x_l$.
\begin{align}
\pi_{l,j,k,+}(x) &= (\theta_{l,j} + v\varphi_k)x \\
\pi_{l,j,k,-}(x) &= (\theta_{l,j} - v\varphi_k)x \\
\pi_{l,j}(x) &= \theta_{l,j}x
\end{align}
   \STATE for $k \in \{1, 2,\dots,N\}$.
   \STATE Sort the directions $k$ by $\max\{r(\pi_{l,j,k,+}),r(\pi_{l,j,k,-})\}$, denote by $\varphi_{k}$ the $k$-th largest direction, and by $r(\pi_{l,j,k,+})$ and $r(\pi_{l,j,k,-})$, the corresponding rewards.
   \STATE{Update the parameters $\theta_l$ of policy $\pi_l$ using the top-$b$ performing rollouts (with $\theta_{l,0}=0$)
\begin{align}
\theta_{l,j+1} = \theta_{l,j} + \frac{\alpha}{b\sigma_R} \sum_{k=1}^{b} \big( r(\pi_{l,j,k,+}) - r(\pi_{l,j,k,-}) \big) \varphi_{k}\label{eq:update}
\end{align}
   \STATE where $\sigma_R$ is the standard deviation of the rewards used in the update step.}
   \ENDFOR
   \UNTIL {ARS termination condition is satisfied}
   \STATE Perform rollouts for all initial states $x_l \in P_{feas}$ using policy $\pi_l$. For each $\pi_l$ select the rollout yielding the highest reward and evaluate using $f(x)$ and extend the prior with the new evaluation
   \ENDFOR
\end{algorithmic}
\end{algorithm}
\end{figure}

\section{Method}
\label{section:method}

In this work, we solve a multi-objective HPO problem using a combination of Bayesian optimization and RL. Unlike other Bayesian optimization algorithms, our methodology iteratively finds new sets of hyperparameters using an ensemble of competing multi-layer perceptron (MLP)-based RL agents, in particular ARS~\citep{mania2018ars}, to optimize the acquisition function. The RL agents operate on the hyperparmaeters as their action space. Our policies only use two linear layers with a hidden layer size of 64 resulting in a couple of thousand trainable parameters each. 
ARS is an RL agent that learns a set of linear (or MLP) policies $\pi_\theta$, parameterized by a set of vectors $\theta$, one for each layer, to control a dynamic environment $\E_\xi$, with $\xi$ encoding the randomness of the environment, maximizing/minimizing an average reward/loss $r(\pi_\theta, \xi)$. ARS achieves this by optimizing over the set of parameters $\theta$ by utilizing derivative-free optimization with noisy function evaluations and for each layer iteratively performing updates using directions of the form
\(\frac{r(\pi_{\theta+\nu\varphi},\xi_1) - r(\pi_{\theta-\nu\varphi},\xi_2)}{\nu}\)
for two i.i.d. random variables $\xi_1$ and $\xi_2$, $\nu$ a positive real number ($0.008$ in our case), and $\varphi$ a zero mean Gaussian vector with i.i.d. standard normal entries. The two rewards $r(\pi_{\theta+\nu\varphi},\xi_1)$ and $r(\pi_{\theta-\nu\varphi},\xi_2)$ are obtained by collecting two sets of trajectory rollouts from the system of interest.

We initialize the training of our competing ARS agents with different samples from the current feasible Pareto front $\mathcal{P}_{\text{\textit{feas}}}(X)$. To find the most diverse set of samples, we perform k-means clustering on the Pareto front if it consists of more elements than we can train policies. To query the objective values, i.e., accuracy, memory, and FLOPS, for a proposed set of hyperparameters of a given DNN architecture, our objective function evaluator first trains, prunes, and quantizes the DNN. Afterwards, the DNN is converted into source code using an automatic code generator to accurately assess the memory usage of the weights and activation tensors. Our implementation performs this procedure using the compression and deployment pipeline proposed by \citet{deutel_deployment_2022}, who describe a configurable pipeline that supports both DNN pruning and full weight quantization. Following their results, we focus our experiments on iterative filter pruning and post-training static quantization.

We integrated ARS into MOBOpt as a parameterized trainable solver, which can thus be seen as an alternative to traditional Bayesian solvers, see Algorithm~\ref{alg:nn_ars}.
The derivative-free random sampling approach implemented by ARS is computationally inexpensive and highly parallelizable, allowing efficient GPU-accelerated execution.

We define the states $x_0, x_1, \dots, x_L$ of our $L$ competing agents as vectors of the parameters of the search space of our optimization problem. Each agent performs actions based on its trainable policy $\pi_\theta$ by altering the parameter values of its state vector within given bounds.
We train the competing policies of our ARS agents for each sample after evaluating the objective function by performing rollouts on posteriors sampled from the GP. We fit them by maximizing the marginal likelihood of the previously queried evaluation of the objective function (see also \citet{rasmussen_gp_for_reg}). The policy training performed for each sample ends after a maximum number of training steps have been conducted. To obtain the initial prior for the GP, we propose candidates during the first ten iterations of optimization using Latin Hypercube sampling~\citep{mckay2000comparison}. We chose this approach over random sampling because we found that by using Latin Hypercube sampling, we could significantly improve the expressiveness of our priors and, therefore, enhance the performance of all evaluated GP-based sampling strategies.

As a reward $r(x)$, we first scale our $n=4$ objectives (accuracy, ROM/RAM memory, FLOPS) to a single objective value using augmented Chebyshev scalarization~\citep{knowles2006parego}, see Eq.~\eqref{eq:scale}, where $\rho = 0.005$ is a small positive value and $\lambda$ is a weight vector that we draw uniformly at random at the beginning of each sample. We then calculate the difference between the scaled objective and the best previously visited (scaled) objective $f^*$ as the reward, see Eq.~\eqref{eq:reward}.
\begin{align}
f_\lambda(x)&=\max_{j=1}^{n}(\lambda_{j}f_j(x)) + \rho\sum_{j=1}^{n}\lambda_{j}f_j(x) \label{eq:scale}\\
r(x)&=
\begin{cases}
    f_\lambda(x) - f_\lambda(f^*),&\text{if } (f_\lambda(x) - f_\lambda(f^*)) > 0\\
    1e^{-3}(f_\lambda(x) - f_\lambda(f^*)),&\text{otherwise}
\end{cases}\label{eq:reward}
\end{align}
%
To avoid having to compute the expected value over the posterior of the GP when computing the acquisition function, a common approach is to use Monte-Carlo (MC) sampling as an approximation to evaluating the integral over the posterior distribution. We implement a quasi-MC sampling approach to estimate the reward used to train the ARS agents with a fixed set of base samples.

As a result, the reward $r(x)$ used by our ARS agents is comparable to the acquisition functions used in traditional Bayesian optimization. In particular, our approach is similar to the expected improvement (EI) used by ParEgo~\citep{knowles2006parego}, with the notable difference that we consider penalized "negative improvements" instead of clamping them to zero. The main reason for clamping "negative improvement" in combination with a Monte Carlo-based evaluation of the GPs is to increase the attractiveness of regions with high uncertainty, and is therefore the mechanism of EI to control the tradeoff between exploration and exploitation. However, we argue that we can still achieve the desired tradeoff if we instead penalize "negative improvements" by multiplication with a small constant factor, while allowing our ARS policies to learn from situations where the current objective function evaluation is worse than the best previously visited one.

After training the competing ARS agents, we perform several rollouts for each of them starting from their respective starting points selected from $\mathcal{P}_{\text{\textit{feas}}}(X)$. We then select the policy that yielded the overall best reward to propose the next set of hyperparameters to be evaluated on the objective function in the next sample. A variant of this approach would be to consider a set of best-performing parameterizations as candidates for a batched optimization in different segments of the Pareto front.
\section{Evaluation}
\label{section:evaluation}

\begin{table}[t]
    \centering
    \caption{Hyperparameters considered by the multi-objective optimization. Learning rate schedule, pruning start, and end are relative to the epochs.}
    \label{tab:hyperparam}
    \begin{tabular}{l c c}
    \toprule
    \textbf{Parameter} & \textbf{Search Space} & \textbf{Interval} \\
    \midrule
    epochs & uniform & $[100, 500]$\\
    batch size & uniform & $[20, 200]$\\
    learning rate (lr) & log. uniform & $[1e^{-5}, 1e^{-2}]$\\
    momentum & log. uniform & $[0.7, 0.99]$\\
    lr schedule & uniform & $[0.4, 0.9]$\\
    lr gamma & uniform & $[0.4, 0.9]$\\
    weight decay & log. uniform & $[0.6, 0.99]$\\
    \midrule
    pruning start & uniform & $[0.0, 0.6]$\\
    pruning end & uniform & $[0.8, 0.95]$\\
    pruning steps & uniform & $[1, 20]$\\
    pruning sparsity & uniform & $[0.1, 0.99]$\\
    \bottomrule
    \end{tabular}
\end{table}

We applied our multi-objective DNN hyperparameter optimization to two use cases with 150 samples each: (1) image classification (32x32 RGB images from CIFAR10~\citep{krizhevsky_learning_2009}) using a reduced version of ResNet18~\citep{he2016resnet} (three residual blocks instead of four, 1.6M initial parameters) and (2) time-series classification of (daily) human activities (DaLiAc)~\citep{leutheuser2013hierarchical} with a window length of 1024 datapoints, using a down-scaled version of MobileNetv3~\cite{howard2019searching} (2.8M initial parameters). Compared to models commonly used by relevant benchmarks for embedded AI, e.g., TinyML Perf~\citep{banbury2021mlperf}, we did not reduce the initial numbers parameters as our optimization process uses pruning and therefore dynamically adjusts the exact number of neurons during training. We trained our models using Stochastic Gradient Decent (SGD) with an exponential learning rate decay. For the optimization, we chose 256 KB of RAM, 1 MB of ROM and $1e^{9}$ FLOPS as constraints imposed by common microcontroller platforms.

The detailed list of hyperparameters considered by our optimizer and their search space intervals derived from expert knowledge can be found in Table~\ref{tab:hyperparam}. These parameters also define the action space of our ARS agents. We consider seven parameters that control general properties of DNN training (training length, batch size, optimizer properties, and learning rate) and four that control filter pruning. We use the automated gradual pruning algorithm~\citep{zhu_prune_2017} to generate our pruning schedule. t takes as input a start and end epoch calculated relative to the number of training epochs and a number of iteration steps. Our optimization considers the pruning sparsity hyperparameter, i.e., how many filters are removed, for each convolutional layer separately. Therefore, the search space considered increases with the depth of the optimized DNN as for each prunable layer an additional pruning sparsity parameter is added which means that the search space is extended by one dimension.

\subsection{Algorithmic Benchmark}
\label{sec:algobench}

\begin{figure}[t]
    \centering
        \includegraphics[width=0.3\textwidth]{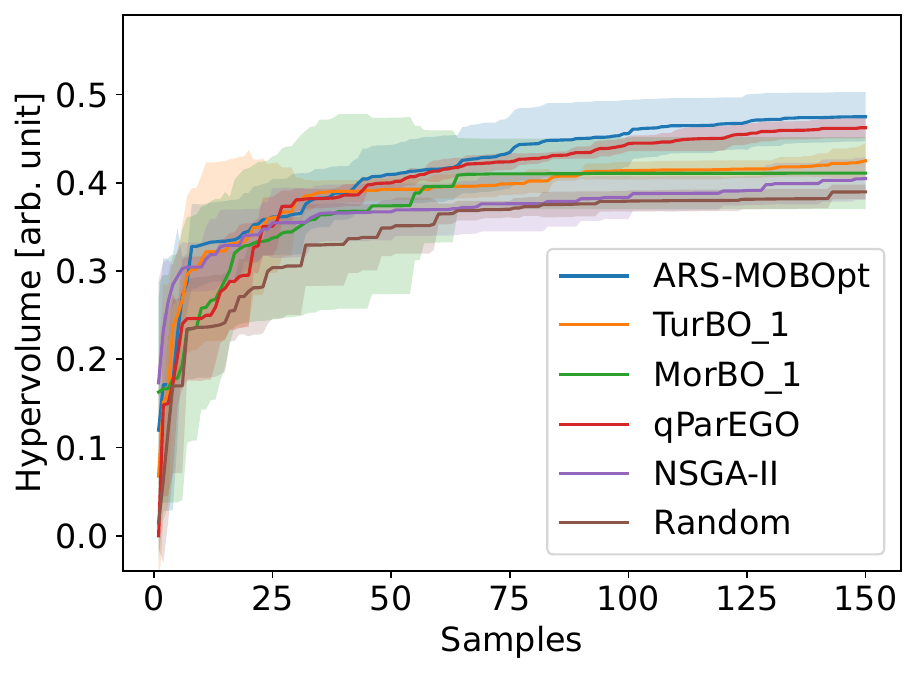}
        \includegraphics[width=0.69\textwidth]{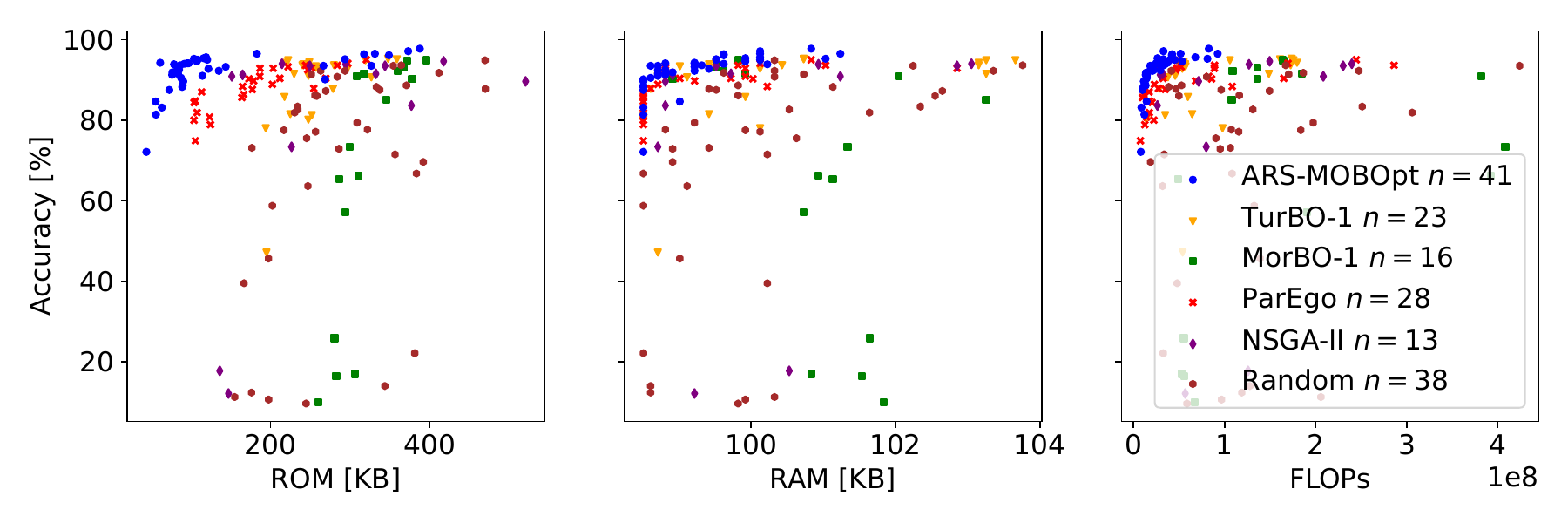}\\
        \includegraphics[width=0.3\textwidth]{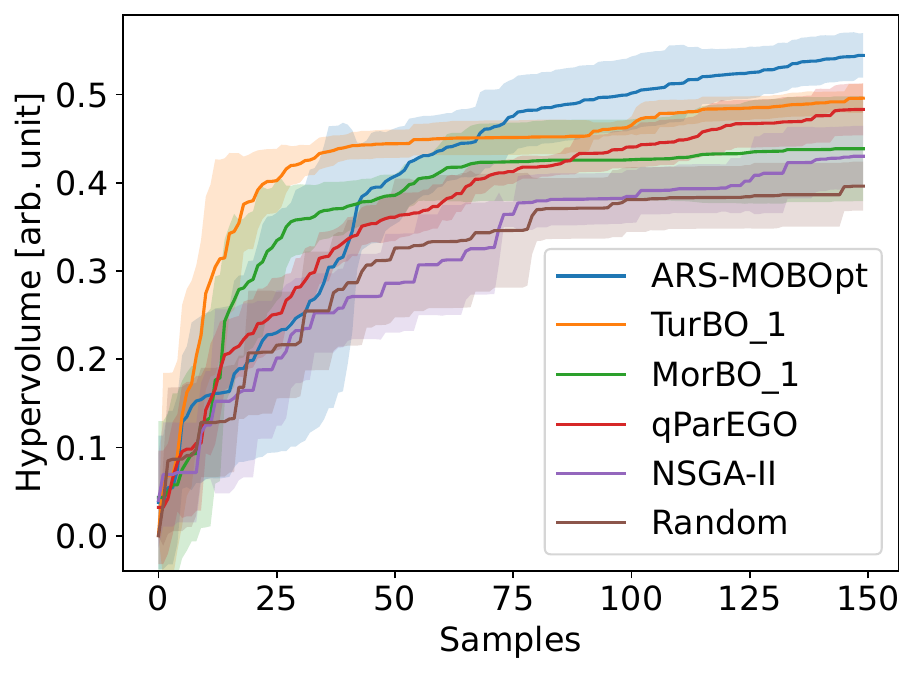}
        \includegraphics[width=0.69\textwidth]{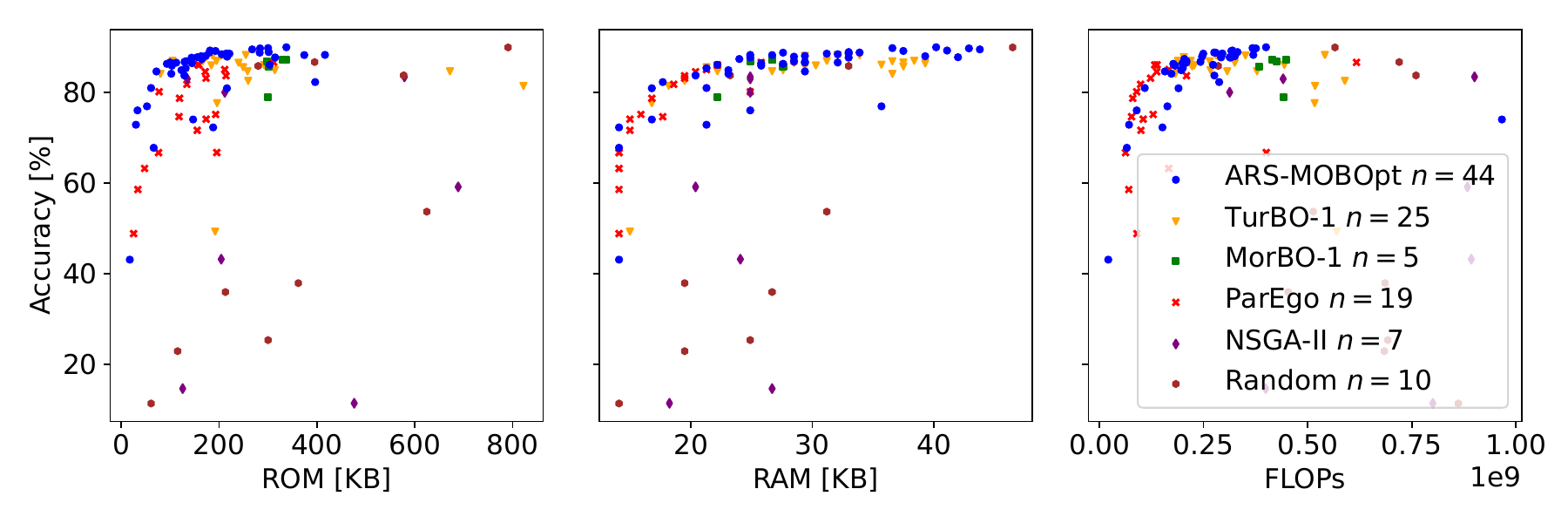}
    \caption{ARS-MOBOpt compared to several baseline approaches (top row: MobileNetv3 on DaLiAc; bottom row: ResNet on CIFAR10). Left column: Our approach (ARS-MOBOpt) outperforming all others in terms of Hypervolume after 50 to 75 evaluated samples. Remaining columns: feasible Pareto sets determined by ARS-MOBOpt compared to other Bayesian (ParEGO, TurBO, MorBO) and evolutionary (NSGA-II) approaches with random sampling (Random) as a baseline.}
    \Description{ARS-MOBOpt compared to five baseline approaches TurBO, MorBO, qParEGO, NSGA-II, and random sampling. The figure has two rows and two columns of plots. The top row shows results for MobileNetv3 on DaLiAc, and the bottom row shows results for ResNet on CIFAR10. The plots in the left column show that our approach (ARS-MOBOpt) outperforms all other baselines in terms of Hypervolume after 50 to 75 evaluated samples. The plots in the right column show the feasible Pareto sets determined by ARS-MOBOpt compared to the baselines.}
    \label{fig:algorithmic}
\end{figure}

As a first experiment, we compared the quality of our approach (ARS-MOBOpt) with several well-known Bayesian optimization strategies (ParEGO~\citep{knowles2006parego}, TurBO~\citep{eriksson2019turbo}, MorBO~\citep{daulton2022multiobjective}), one evolutionary approach (NSGA-II~\citep{deb2002nsgaii}), and a random sampling approach as a baseline, see Fig.~\ref{fig:algorithmic}. All Bayesian optimization strategies including our novel ARS-based BO strategy were implemented with BoTorch~\citep{balandat2020botorch}, which we used in conjunction with Optuna~\citep{akiba_optuna_2019} as a metaheuristic optimization framework. For NSGA-II and Random we used the implementations provided by Optuna.

We gave each optimization strategy a search budget of 150 samples to evaluate the objective functions and monitored the improvement of the Hypervolume over the samples. Higher Hypervolume values are considered superior. For all tested Bayesian strategies the first 10 samples were conducted as priors using Latin Hypercube sampling~\citep{mckay2000comparison}. We normalized our objectives accuracy, ROM, RAM, and FLOPs to their respective constraints and used the maximum feasible objective value as a common reference when computing the Hypervolume. We performed five independent seeds for each strategy and provide the observed variance.

Fig.~\ref{fig:algorithmic} shows the results for two common DNN architectures, MobileNetv3 and ResNet18, which we trained using two datasets, DaLiAc and CIFAR10. The plots show the observed Hypervolume over the number of optimization samples. Therefore, the resulting curves, one for each optimization strategy, describe the improvement of the Hypervolume. First, all informed solvers outperformed the independent random sampling baseline on both use cases. Second, all the approaches that are based on Bayesian optimization, i.e., TurBO, MorBO, ParEGO and ARS-MPBOpt, which consider the constraints of a very tight search budget of 150 samples, performed better than the evolutionary sampling strategy (NSGA-II). Additionally, for both use cases, a significant Hypervolume improvement is observed when using our ARS-based sampling strategy with 3000 sampling directions, a top-$k$ selection percentage of 1\%, and a rollout horizon of 4 to train the competing policies. This implies that by using a trainable RL-based strategy to determine new promising parametrizations as the optimization progresses, our algorithm makes much better use of the knowledge about the search space encoded in the surrogate model than the other non-trainable Bayesian sampling approaches. 

In addition, we provide qualitative results for both considered use cases, see the remaining columns in Fig.~\ref{fig:algorithmic}. The plots show the Pareto optimal samples for our approach (in blue) compared to all other algorithms tested on MobileNetv3/DaLiAc (top row) and Resnet/CIFAR10 (bottom row). For both use cases, ARS-MOBOpt was able to find samples that are Pareto-dominant over those proposed by the other algorithms, while satisfying all imposed target constraints ($<$1MB ROM, $<$256KB RAM, $<$1e9 FLOPs). Moreover, for both use cases, the number of elements $n$ in the two sets of feasible Pareto optimal configurations resulting from our approach are larger than the sets found by the other approaches, implying that our algorithm was able to provide a larger variety of Pareto optimal deployable DNNs.

To gain more insight into the overall performance of ARS, we tested it on a subset of the multivariate datasets from the UEA \& UCR time series classification benchmark~\citep{bagnall16bakeoff}. We compare our method with ParEGO using the same setup as described above, see Table~\ref{tab:uea_ucr_small} for all datasets on which reasonable models with an accuracy of at least 70\% are found (and Table~\ref{tab:uea_ucr} in Appendix~\ref{app:UEA-complete} for a complete list). To provide context, we also list reference results for InceptionTime~\citep{fawaz2020inception} that we extracted from \citet{bagnall16bakeoff}. We let both ARS and ParEGO optimize a CNN architecture with a single regular convolution followed by 10 depth-wise separable convolutions with batchnorm and with a total of 5 squeeze and excitation blocks~\citep{hu2018squeeze} after convolutions 10, 8, 6, 4, and 2, followed by adaptive average pooling and a single linear layer for classification (2.6m initial parameters). Our approach achieves a better Hypervolume than ParEGO in 7 of 10 datasets (and is en par on one dataset). If we also consider datasets where we found models with predictive accuracies below 70\% our approach outperforms ParEGO in almost two-thirds of the datasets, see Appendix~\ref{app:UEA-complete}. The results also show that, compared to ParEGO, our algorithm was generally better at finding DNNs with higher accuracy scores, but often at the cost of higher memory and FLOPs requirements.

\begin{table*}[t]
    \centering
    \caption{Excerpt from the results of the multivariate datasets of the UEA \& UCR time series classification benchmark~\citep{bagnall16bakeoff} for our ARS strategy (columns 3--7) and ParEGO (columns 8--12). We show the achieved Hypervolume after 150 trials as well as the maximum accuracy. We also show the minimum memory and FLOPs required to achieve at least 70\% accuracy.}
    \label{tab:uea_ucr_small}
    \resizebox{\textwidth}{!}{
    \begin{tabular}{l c c c c c c c c c c c}
\toprule
\textbf{Dataset} & \color{lightgray}{\textbf{InceptionTime}} & \multicolumn{5}{c}{\textbf{ARS}} & \multicolumn{5}{c}{\textbf{ParEGO}} \\
& \color{lightgray}{Acc. [\%]} & HV & Acc. [\%] & ROM [Kb] & RAM [Kb] & FLOPs & HV & Acc. [\%] & ROM [Kb] & RAM [Kb] & FLOPs \\
\midrule
Epilepsy & \color{lightgray}{98.55} & \textbf{0.97} & \textbf{97.81} & \num{120512} & \num{4920} & \num{3570578} & 0.93 & 94.16 & \textbf{\num[detect-weight]{66988}} & \textbf{\num[detect-weight]{3392}} & \textbf{\num[detect-weight]{2050650}} \\
BasicMotions & \color{lightgray}{100.00} & 0.96 & \textbf{97.50} & \num{131176} & \num{3000} & \num{2745720} & \textbf{0.97} & \textbf{97.50} & \textbf{\num[detect-weight]{86696}} & \textbf{\num[detect-weight]{2500}} & \textbf{\num[detect-weight]{1493195}} \\
Cricket & \color{lightgray}{98.61} & \textbf{0.96} & \textbf{97.22} & \num{100404} & \textbf{\num[detect-weight]{29924}} & \num{16020432} & 0.95 & 95.83 & \textbf{\num[detect-weight]{76068}} & \num{30520} & \textbf{\num[detect-weight]{10139994}} \\
ERing & \color{lightgray}{87.78} & \textbf{0.96} & \textbf{96.67} & \num{100256} & \num{1988} & \num{2025968} & 0.93 & 93.33 & \textbf{\num[detect-weight]{73196}} & \textbf{\num[detect-weight]{1228}} & \textbf{\num[detect-weight]{1002400}} \\
ArticularyWordRecognition & \color{lightgray}{98.33} & \textbf{0.94} & \textbf{95.64} & \num{150516} & \num{6180} & \num{4194267} & 0.92 & 92.36 & \textbf{\num[detect-weight]{80748}} & \textbf{\num[detect-weight]{5328}} & \textbf{\num[detect-weight]{2127501}} \\
UWaveGestureLibrary & \color{lightgray}{87.81} & \textbf{0.91} & \textbf{91.67} & \num{110256} & \num{6944} & \num{5779512} & \textbf{0.91} & \textbf{91.67} & \textbf{\num[detect-weight]{72052}} & \textbf{\num[detect-weight]{4092}} & \textbf{\num[detect-weight]{3027980}} \\
NATOPS & \color{lightgray}{96.11} & \textbf{0.86} & \textbf{87.22} & \num{129032} & \num{5784} & \num{2793175} & 0.78 & 78.33 & \textbf{\num[detect-weight]{100264}} & \textbf{\num[detect-weight]{4944}} & \textbf{\num[detect-weight]{1118016}} \\
SelfRegulationSCP1 & \color{lightgray}{83.96} & \textbf{0.79} & \textbf{79.85} & \num{104688} & \textbf{\num[detect-weight]{22400}} & \num{17814384} & 0.78 & 78.73 & \textbf{\num[detect-weight]{78052}} & \textbf{\num[detect-weight]{22400}} & \textbf{\num[detect-weight]{7147130}} \\
PEMS-SF & \color{lightgray}{--} & 0.76 & 81.50 & \textbf{\num[detect-weight]{137784}} & \textbf{\num[detect-weight]{554832}} & \textbf{\num[detect-weight]{7984536}} & \textbf{0.79} & \textbf{84.97} & \num{238016} & \num{555756} & \num{57178800} \\
Heartbeat & \color{lightgray}{58.05} & \textbf{0.73} & \textbf{74.51} & \num{81648} & \textbf{\num[detect-weight]{99224}} & \num{8139132} & 0.73 & 74.02 & \textbf{\num[detect-weight]{62532}} & \textbf{\num[detect-weight]{99224}} & \textbf{\num[detect-weight]{2793768}} \\
\bottomrule
\end{tabular}}
\end{table*}

\subsection{Robustness and Hyperparameter Selection}
\label{sec:ablation}

\begin{figure}[b]
    \centering
    \begin{subfigure}[t]{.32\textwidth}
        \centering
        \includegraphics[width=\textwidth]{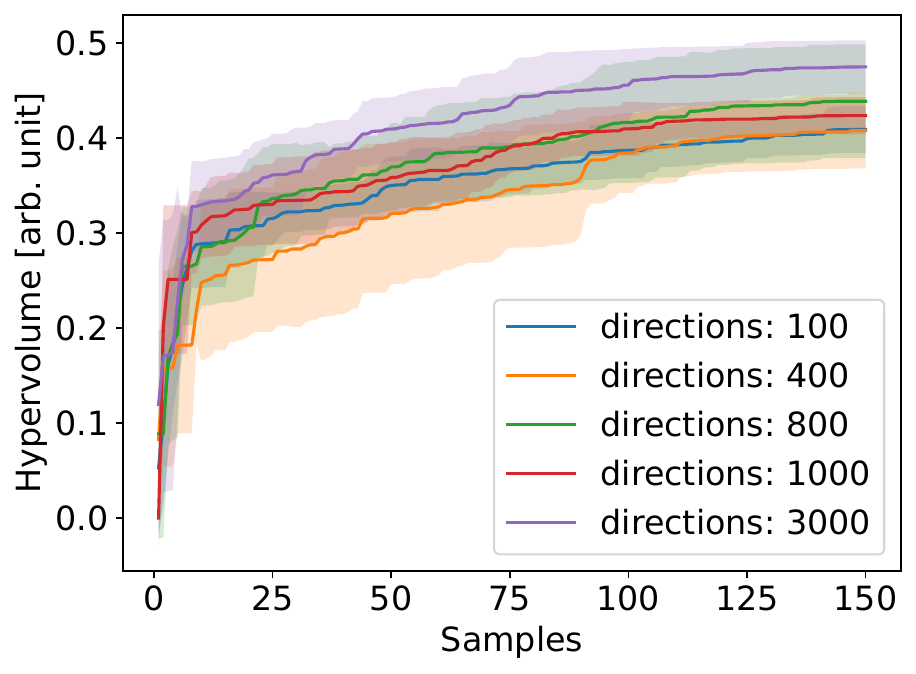}
        \caption{Number of \textit{sampling directions $N$} performed for each step.}
        \label{fig:abl_daliac_directions}
    \end{subfigure}
    \hspace{1mm}
    \begin{subfigure}[t]{.32\textwidth}
        \centering
        \includegraphics[width=\textwidth]{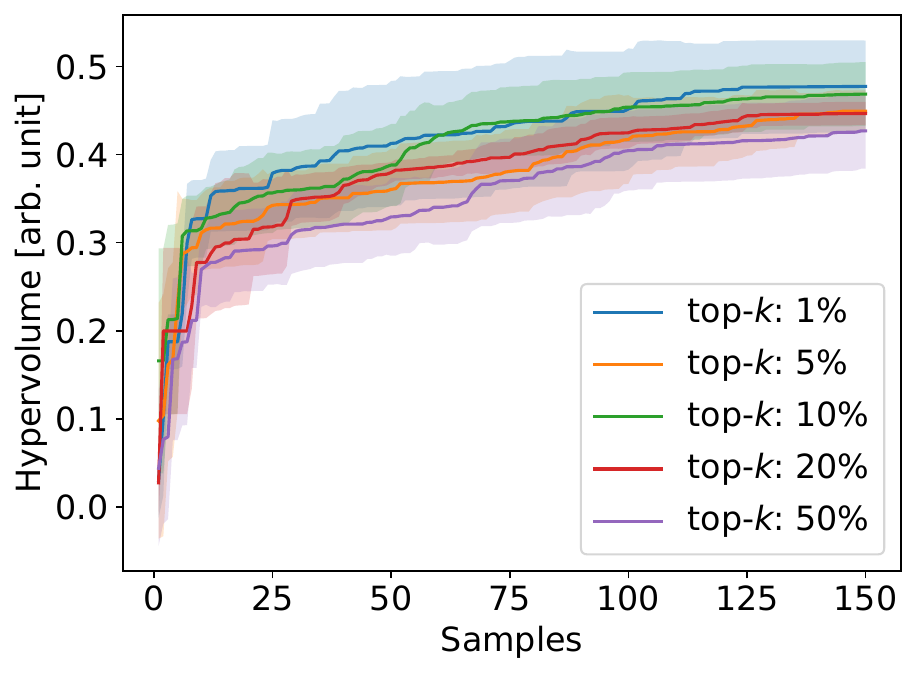}
        \caption{Percentage of \textit{top-$k$} performing samples selected for policy updates.}
        \label{fig:abl_daliac_topk}
    \end{subfigure}
    \hspace{1mm}
    \begin{subfigure}[t]{.32\textwidth}
        \centering
        \includegraphics[width=\textwidth]{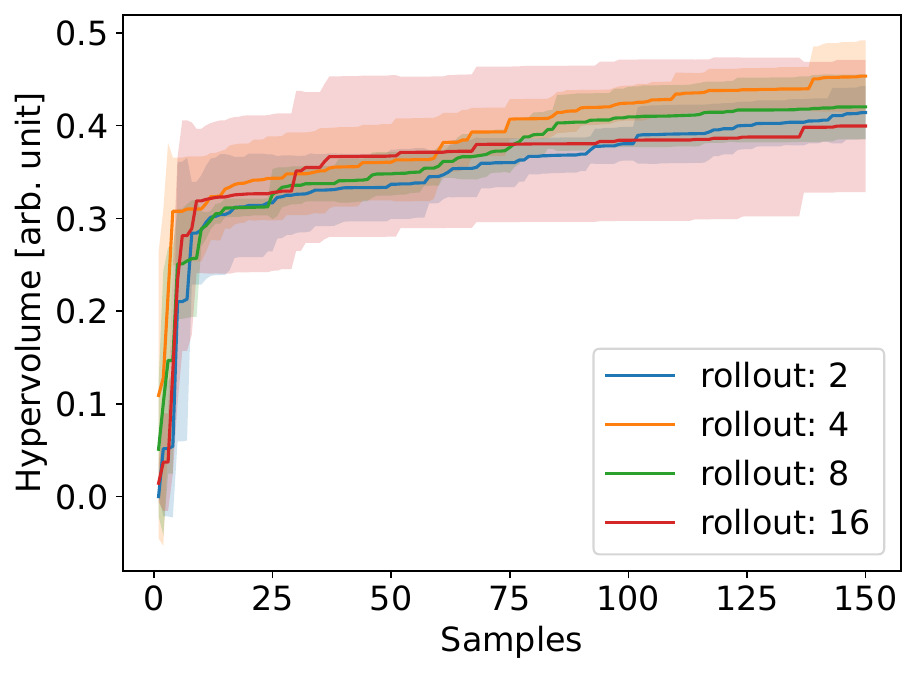}
        \caption{\textit{Rollout Horizon $H$} performed by the competing policies.}
        \label{fig:abl_daliac_rollout}
    \end{subfigure}
    \caption{Detailed results examining the effects of the three key ARS hyperparameters on the achieved Hypervolume, 5 seeds each. MobileNetv3, 1.6M init. params, DaLiAc dataset, window length 1024.}
    \Description{Detailed results examining the effects of the three most important ARS hyperparameters, sampling directions, top-k selected samples, and rollout horizon, on the achieved Hypervolume for MobileNetv3 trained on the DaLiAc dataset with a window length of 1024. The three plots in the figure show that 3000 sample directions, a top-k selection rate of 1\%, and a rollout horizon of 16 resulted in the best Hypervolume improvement over 150 samples.}
    \label{fig:abl}
\end{figure}

Our proposed ARS sampling strategy can be configured with different hyperparameters that affect the behavior of the ARS algorithm, see Alg.~\ref{alg:nn_ars} for a complete listing. The most important are (a) the sampling directions $N$, (b) the top-$k$ selection rate, and (c) the rollout horizon $H$. Both the optimization progress and the quality of explored solutions obtained by our algorithm can vary depending on their parameterization, see Fig.~\ref{fig:abl}: In general, a top-$k$ selection rate of $1\%$, a number of sampling directions around $3000$, and a rollout horizon of $4$ yielded the best results. In the following, we want to further discuss the impact of these key hyperparameters, to provide a better understanding of their effects.

First, especially for complex optimization problems with a high-dimensional parameter space, a high number of sampling directions significantly improved the Hypervolume that our ARS-based sampling strategy was able to achieve, compared to a low number of directions, which often resulted in a below-average performance, see Fig.~\ref{fig:abl_daliac_directions}. However, we found, that it is not feasible to increase the number of directions indefinitely to support more and more complex problems, as this also significantly increases the time required to train the ARS policies and the memory requirements of our algorithm. Instead, it should be noted that the exact number of sampling directions required to obtain a competitive result is highly correlated with the complexity of the underlying optimization problem. 

Second, when increasing the top-$k$ percentage of sampling directions selected by ARS to perform policy updates, we observed that only selecting a small subset of $1\%$ of the sampled directions to perform policy updates improved the Hypervolume achieved by ARS compared to using larger subsets, see Fig.~\ref{fig:abl_daliac_topk}. Our observation is thus consistent with \citet{mania2018ars}'s argument for introducing the hyperparameter as part of ARS. The authors point out that including all observed rewards in the parameter update can be detrimental to performance because outliers can easily introduce bias.

Third, we found that an optimal episode length (rollout horizon) is at $H=4$ steps, see Fig.~\ref{fig:abl_daliac_rollout}. Both significantly shorter and longer episodes resulted in lower average performance. However, we would like to point out that we observed a significantly larger variance for longer episodes compared to shorter episodes. This is consistent with the increased variance in the estimation of total expected return in MC-like algorithms such as ARS~\citep{Sutton1998}. 

\subsection{Discussion}
\label{sec:intuition}

\begin{figure}
    \centering
    \begin{subfigure}[t]{.48\linewidth}
        \includegraphics[width=\linewidth]{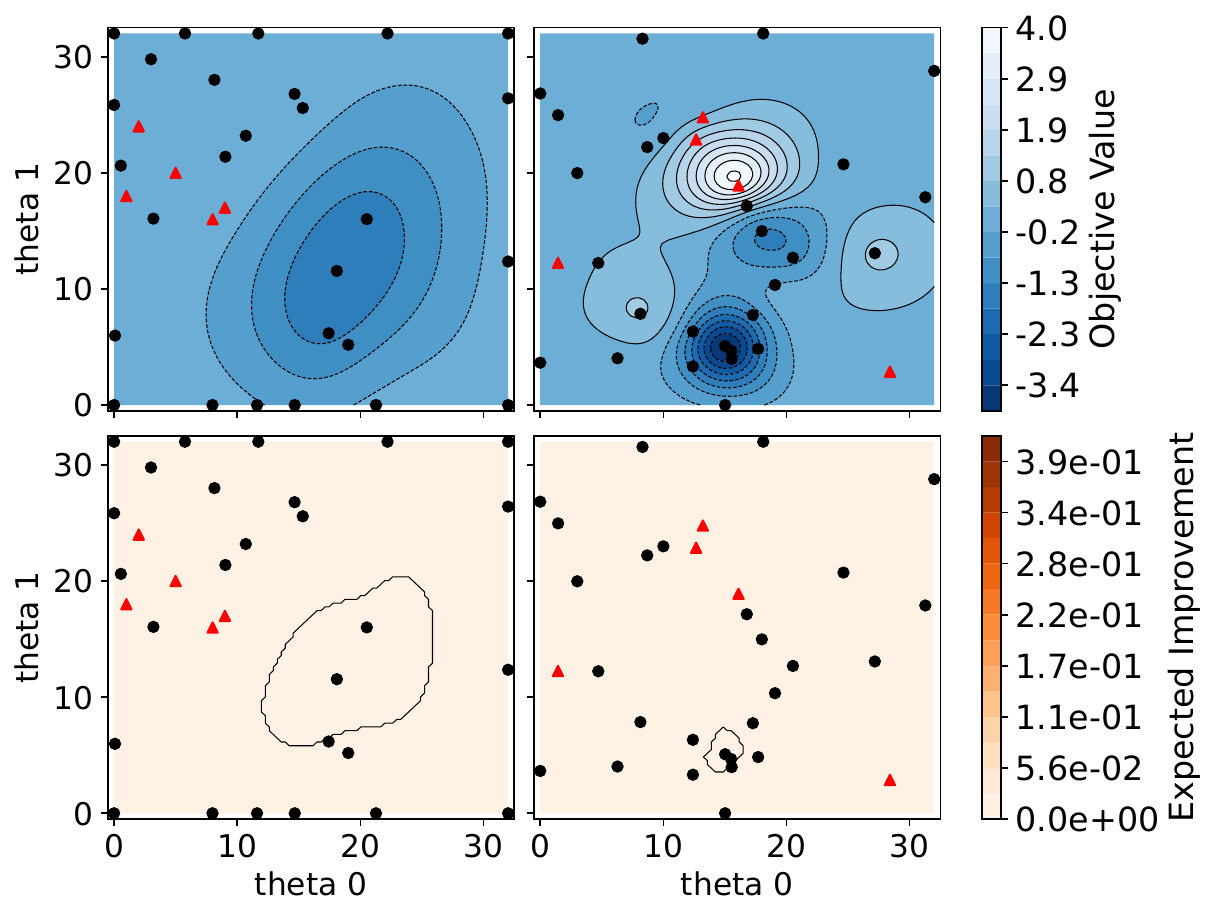}
        \subcaption{ParEGO}
        \label{fig:contour_botorch}
    \end{subfigure}
    \begin{subfigure}[t]{.48\linewidth}
        \includegraphics[width=\linewidth]{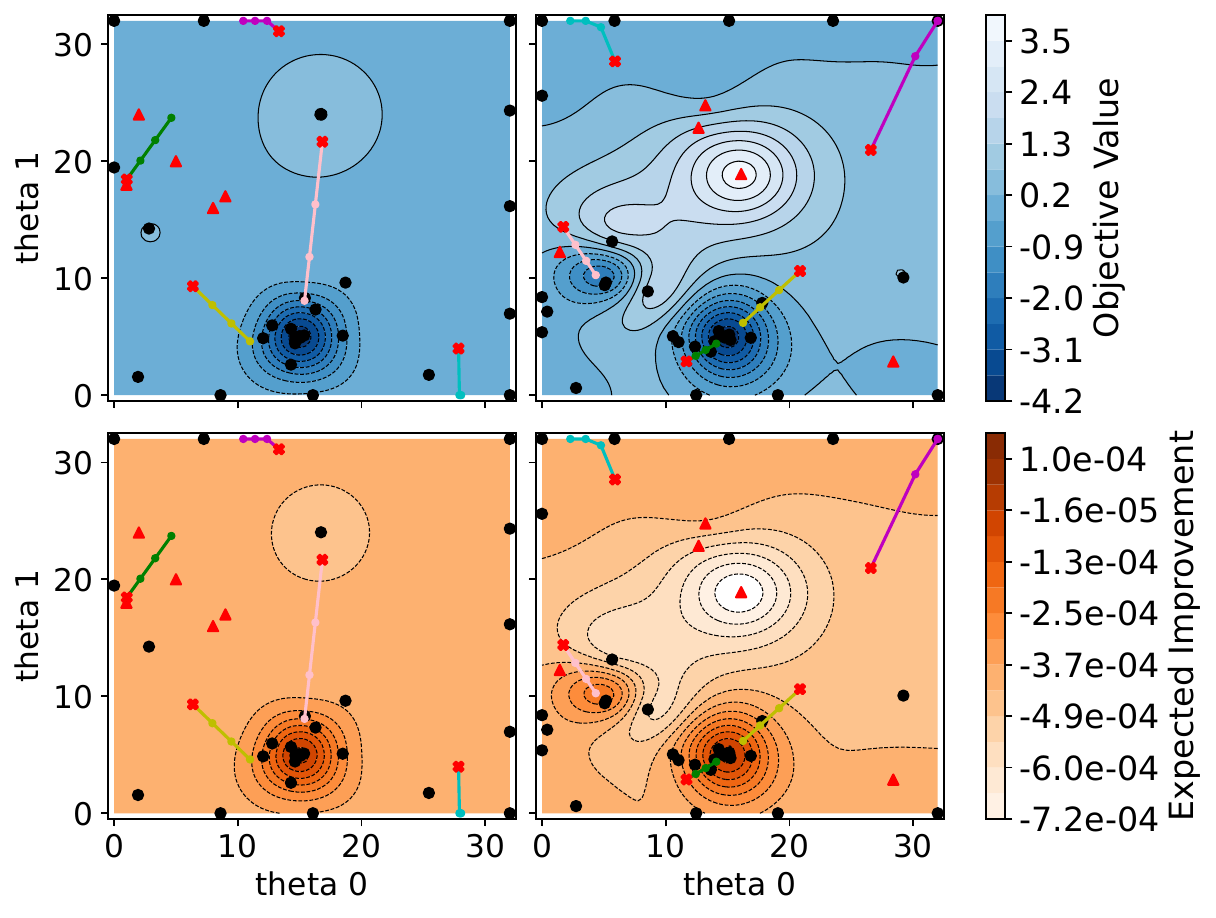}
        \subcaption{ARS}
        \label{fig:contour_ars}
    \end{subfigure}
    \caption{Topography of the optimization landscape (top row) and EI (bottom row) estimated by the Bayesian surrogates for the synthetic optimization problem shown in Fig.~\ref{fig:synth_example} in Appendix~\ref{app:synth_example} after 40 samples for both ParEGO and ARS (ours), given two sets of priors marked with red triangles. The global minimum is at $\theta_0 = 15$, $\theta_1 = 5$. For ARS, the rollouts of the trained policies of the 40th sample are shown as lines with their starting points marked by red crosses.}
    \Description{The topography of the optimization landscape and the EI estimated by the Bayesian surrogates for the synthetic optimization problem after 40 samples for both ParEGO and ARS, given two sets of priors marked by red triangles. For ARS, the rollouts of the trained policies of the 40th sample are shown as lines with their starting points marked by red crosses. The results in the figures show that after 40 samples, ARS was able to sample closer to the global optimum of the synthetic problem than ParEGO. Furthermore, the rollouts of the trained policies show that the ARS agents have learned to navigate the already explored optimization landscape in search of a minimum.}
    \label{fig:synth_example}
\end{figure}

To better understand why our RL-based solver can improve on other Bayesian solvers like ParEgo, we analyzed both using a synthetic optimization problem. Given a two-dimensional single-objective optimization problem with an objective landscape as shown in Fig.~\ref{fig:synth_gt} from ~\citet{passino2005biomimicry} in Appendix~\ref{app:synth_example}. This problem has a challenging landscape with large areas that have little to no gradient combined with some steep minima and maxima. The respective Bayesian surrogates of ParEgo and ARS-MOBOpt as well as the topographies of their expected improvement (EI) after 40 trials are shown in Figs.~\ref{fig:contour_botorch}~and~\ref{fig:contour_ars}. For our approach (ARS-MOBOpt), we also show the rollouts performed during the validation of the local policies for five selected start points (red crosses) after the last (40th) sample. 

We manually selected two sets of priors (marked as red triangles) to provide both algorithms with interesting starting situations: First, a situation where all samples are close together in a region with almost no gradient, see top left in Fig.~\ref{fig:contour_botorch}~and~\ref{fig:contour_ars}. This initialization is interesting because it provides almost no information about the environment, requiring extensive exploration by the algorithms before they can exploit found minima. Second, a more general starting situation with a broader distribution of prior samples, see top right in Fig.~\ref{fig:contour_botorch}~and~\ref{fig:contour_ars}. This initialization focuses more on evaluating how well the algorithms can apply the exploration-exploitation tradeoff and how fast they can find minima. For both priors, we took care not to include samples close to the global minimum.

Looking at the results proposed by ParEgo (marked as black dots) after 40 samples in Fig.~\ref{fig:contour_botorch}, it becomes evident that for both sets of priors the algorithm seems to have a clear focus on exploration, leading to a wide sampling of the search space. However, looking at the two examples, this can lead to situations where even though the global optimum was either not found (left column in Fig.~\ref{fig:contour_botorch}) or it was not fully exploited (right column in Fig.~\ref{fig:contour_botorch}). Furthermore, we argue that this focus on exploration is mainly a result of the definition of the EI as it does not consider any "negative improvement". Looking at the topography of the EI for both examples in the last row of Fig.~\ref{fig:contour_botorch}, it becomes evident that this can result in large parts of the acquisition function having no non-zero gradient. We argue that this is not optimal when relying on gradient-based solvers, as the ParEgo implementation typically does.

In comparison, if we look at the result of our proposed ARS-based approach in Fig.~\ref{fig:contour_ars}, it can be seen that our competing local policies, visualized as lines with red crosses as their starting point, have become experts in solving the environment around their starting point and can therefore provide sound solutions even in regions without significant gradients. Therefore, compared to ParEgo, we observed a more directed exploration and a clearer focus on exploiting environmental knowledge, compared to Fig.~\ref{fig:contour_botorch}. However, despite this more pronounced focus on exploitation, due to the competing ensemble of agents, our solver was still able to consistently escape local minima, e.g. see Fig.~\ref{fig:contour_ars} (right col.).

\subsection{Comparison with Related Optimization Approaches}
\label{sec:compare}

In the following, we provide comparative results of our approach with other RL, NAS, and automated pruning methods. Namely, we analyze the feasibility of using a different type of RL agent, PPO, instead of ARS, and compare our approach to both MCUNet, zero-shot NAS, and automated pruning.

\subsubsection{Comparison with PPO}
\label{sec:ppo}

The following experiment is used to compare the performance of ARS with PPO~\citep{schulman2017proximal}. PPO is an on-policy gradient RL algorithm that, unlike comparable algorithms such as TRPO~\citep{schulman2015trust}, requires only first-order optimization while achieving better overall stability and performance. Because PPO is sample-efficient and can be used for environments with continuous action spaces, we selected it as a suitable candidate for comparison with the ARS agents we have used so far.

\begin{figure}[t]
    \centering
    \begin{subfigure}[t]{.32\textwidth}
        \centering
        \includegraphics[width=\textwidth]{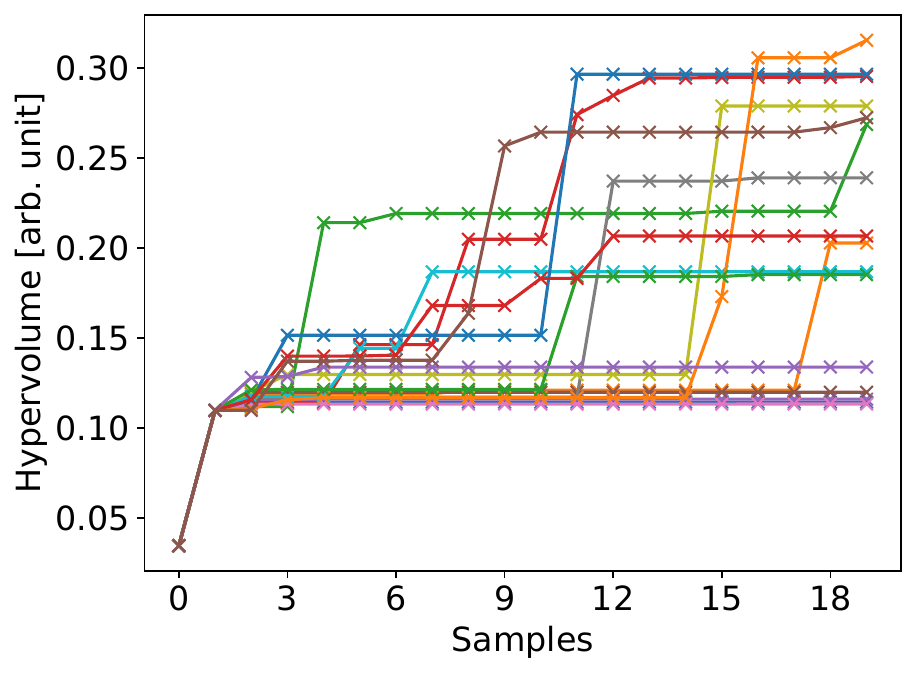}
        \caption{PPO Hyperparameter pre-optimization for ResNet on CIFAR10}
        \label{fig:ppo_hv:study}
    \end{subfigure}
    \begin{subfigure}[t]{.32\textwidth}
        \centering
        \includegraphics[width=\textwidth]{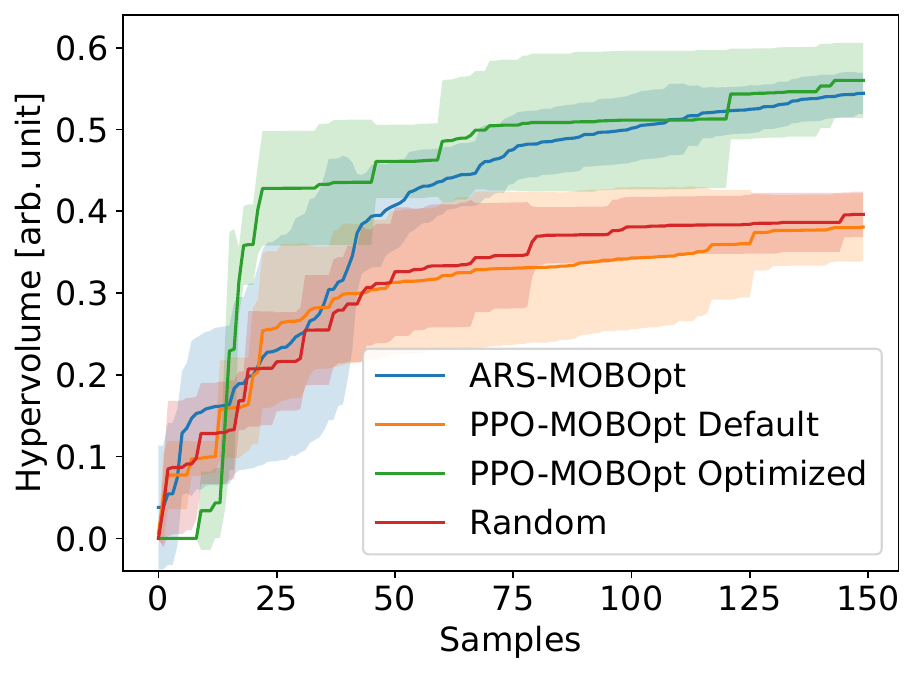}
        \caption{ResNet on CIFAR10}
        \label{fig:ppo_hv:resnet}
    \end{subfigure}
    \begin{subfigure}[t]{.32\textwidth}
        \centering
        \includegraphics[width=\textwidth]{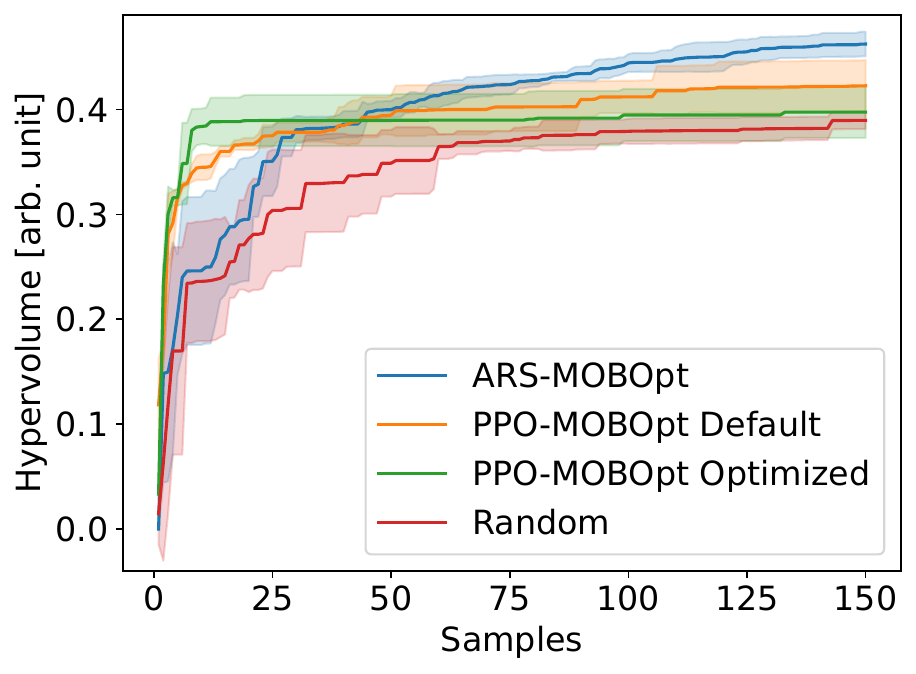}
        \caption{MobileNetv3 on DaLiAc}
        \label{fig:ppo_hv:daliac}
    \end{subfigure}
    \caption{Comparison of MOBOpt with ARS and PPO for ResNet on CIFAR10 and MobileNetv3 on DaLiAc. For both cases we show random sampling as the baseline. The results for ARS and Random are the same as the results in Fig.~\ref{fig:algorithmic}. Since the default hyperparameters of PPO did not produce results of quality better than random sampling for ResNet, we performed a hyperparameter pre-optimization over 20 trials, see (a). The Hypervolume results are shown in (b) for ARS-MOBOpt and pre-optimized PPO. (c) shows the need of problem-specific  pre-optimization of hyperparameters for PPO. The green curve is obtained when applying the pre-optimized parameter settings from (a) to MobileNetv3  rather than ResNet.}
    \Description{Comparison of MOBOpt with ARS and PPO for ResNet on CIFAR10 and MobileNetv3 on DaLiAc. For both cases, random sampling is shown as the baseline. In (a), it can be seen that different PPO hyperparameters resulted in a high variance of the Hypervolume in the pre-optimization for ResNet on CIFAR10. In (b), we show that using the default hyperparameters of PPO results in a Hypervolume that is no better than random sampling, while PPO with optimized hyperparameters from the pre-optimization in (a) results in a Hypervolume that is as good as when using ARS. (c) shows that using PPO with the same hyperparameters optimized for ResNet on CIFAR10 in (a) does not work for MobileNetv3 on DaLiAc, i.e. the achieved Hypervolume is not better than random sampling, demonstrating the need for problem-specific pre-optimization of hyperparameters for PPO.}
    \label{fig:ppo_hv}
\end{figure}

We repeated the optimization of ResNet on CIFAR10 and MobileNetv3 on DaLiAc as described in Section~\ref{sec:algobench} with PPO instead of ARS and provide results in Fig.~\ref{fig:ppo_hv}. For context and comparison, we also show the results for ARS and random sampling which we took from Fig.~\ref{fig:algorithmic}. For all experiments, we used the stablebaselines~\citep{raffin2021stable} implementation of PPO and repeated each 5 times, recording means and standard deviations as we did with ARS and random sampling in Section~\ref{sec:algobench}.

Initially, we tested PPO with its default parameters. However, this resulted in a worse Hypervolume improvement than random sampling, see the orange and red curves in Fig.~\ref{fig:ppo_hv:resnet}. Furthermore, when analyzing the orange curve, it can be observed that during the first part of the optimization PPO performed well in efficiently exploring the search space, resulting in a roughly similar Hypervolume compared to ARS after about 40 samples. However, the improvement started to stagnate sharply after this initial exploration-heavy phase. As a result, we found that after the initial exploration phase, all PPO agents started to have a strong tendency to get trapped in local optima found by previous samples, which resulted in the agents completely stopping their exploration.

Considering the above results, we performed a PPO hyperparameter optimization using Optuna~\citep{akiba_optuna_2019} for the ResNet on CIFAR10 optimization problem, see Fig.~\ref{fig:ppo_hv:study}. For each PPO configuration tested, we ran MOBOpt on 20 samples with a fixed prior of 2 samples. A complete listing of the search space can be found in Appendix~\ref{app:ppo_params}. The results presented in Fig.~\ref{fig:ppo_hv:study} show that simply changing the hyperparameters of PPO, while leaving the MOBOpt problem as it is, led to very different Hypervolume improvements, i.e., drastically different behaviours of the PPO agents. Most notably, we observed that for many configurations, PPO was not able to improve on the prior at all, while for other configurations, the algorithm achieved a high Hypervolume and a significant improvement over the prior.

Using the best performing set of PPO hyperparameters from the optimization shown in Fig.~\ref{fig:ppo_hv:study}, we repeated the experiment for ResNet on CIFAR10, see the green curve in Fig.~\ref{fig:ppo_hv:resnet}. It can be observed that by using the optimized PPO configuration the algorithm achieved a Hypervolume improvement that consistently even exceeded the one we observed for ARS. Moreover, especially during the first 40 samples, PPO was able to improve the Hypervolume much faster than any of the other algorithms we tested.

Finally, we tested how well the PPO hyperparameters optimized for ResNet on CIFAR10 abstracts to another related optimization problem. Therefore, we repeated the second experiment we discussed earlier in Fig.~\ref{fig:algorithmic}, i.e. MobileNetv3 on DaLiAc, with PPO and the hyperparameters optimized for ResNet, see Fig.~\ref{fig:ppo_hv:daliac}. The results of this additional experiment show that while PPO was again initially able to reach a better Hypervolume faster than all other algorithms we tested, after the first 20 samples the improvement sharply stagnated and the algorithm stopped finding significantly better points. This is similar to the observations we made when testing PPO for ResNet on CIFAR10 with default parameters. As a result, with the hyperparameters from Fig.~\ref{fig:ppo_hv:study} optimized for ResNet, PPO again barely outperformed the random sampling baseline for the MobileNetv3 on DaLiAc optimization problem, and even performed worse than when using its default parameters, compare green and orange curves in Fig.~\ref{fig:ppo_hv:daliac}.

Overall, the results we obtained align with common observations for deep reinforcement learning algorithms like PPO, where optimal hyperparameters can vary significantly across tasks, often necessitating extensive tuning for new tasks. We have shown that the performance of PPO and its exploration-exploitation tradeoff can be significantly improved by performing a full study of its hyperparameters to adapt them as best as possible to a specific optimization problem. However, especially compared to ARS, where we have discussed the effects of its three hyperparameters on optimization in detail in Section~\ref{sec:ablation}, and where we have shown that they work consistently well for a wide range of tasks, PPO has significantly more complex hyperparameters whose effects are not intuitive, and which we have shown to behave differently for different MOBOpt problems. We conclude that an effective use of PPO would necessitate a full optimization of its hyperparameters for each optimization problem, making it an extremely time-consuming and resource-intensive optimizer for MOBOpt. We emphasize this point because one of the main reasons we chose ARS in the first place was its simplicity.

\subsubsection{Comparison with MCUNet and Zero-Shot NAS}
\label{sec:mcu}

We present a comparison between our results from Fig.~\ref{fig:algorithmic}, five DNN candidates from the MCUNet model list~\citep{lin_mcunet_2020,lin2021mcunetv2}, which are optimized for ImageNet and which we retrained for CIFAR10, and a version of MCUNet that we optimized specifically for CIFAR10 using ProxylessNAS~\citep{cai2018proxylessnas}, see Table~\ref{tab:mcunet}. The original input resolution of the CIFAR10 dataset is 32x32 which is used by our models. For the MCUNet models taken from the model list we report different input resolutions as they were predetermined. For ProxylessNAS, 48x48 was the smallest possible resolution that did not crash the framework. A comparison with the other time series datasets we present in our evaluation is not easily possible, since both MCUNet and ProxylessNAS do not support input with only one spatial dimension. Since our tool reports FLOPs instead of MACs, we assume that one MAC operation (multiply-accumulate) equals two FLOPs, i.e., a multiplication followed by an addition. Since MCUNet uses a different mapping tool than we do, we also report the SRAM and Flash requirements of the mapping achieved by our tool in parentheses.

\begin{table}[t]
\centering
\caption{Comparative results of MCUNet and our approach for CIFAR10. MCUNet-in0-4 are taken from the MCUNet model list, while mcunet-proxyless was optimized specifically for CIFAR10 using ProxylessNAS.}
\label{tab:mcunet}
\resizebox{.7\textwidth}{!}{
\begin{tabular}{llllll}
\toprule
\textbf{Model} & \textbf{MACs} & \textbf{SRAM/RAM} & \textbf{Flash/ROM} & \textbf{Top1 (int8)} & \textbf{Input} \\
\midrule
mcunet-in0          & 6.4M                     & 266KB (60KB)  & 889KB (573KB)   & 80.79\%                     & 48x48px          \\
mcunet-in1          & 12.8M                    & 307KB (147KB) & 992KB (587KB)   & 82.55\%                     & 96x96px          \\
mcunet-in2          & 67.3M                    & 242KB (410KB) & 878KB (586KB)   & 86.24\%                     & 160x160px        \\
mcunet-in3          & 81.8M                    & 293KB (495KB) & 897KB (594KB)   & 86.13\%                     & 176x176px        \\
mcunet-in4          & 125.9M                   & 456KB (614KB) & 1876KB (1437KB) & 87.60\%                     & 160x160px        \\
mcunet-proxyless & 25.5M & 971KB (83KB)  & 2923KB (2842KB) & 82.33\% & 48x48px          \\
ours-small          & 37M                      & 21KB          & 43KB            & 81.00\%                     & 32x32px          \\
ours-medium         & 54M                      & 24KB          & 106KB           & 85.00\%                     & 32x32px          \\
ours-large          & 81.9M                    & 276KB         & 212KB           & 87.96\%                     & 32x32px         \\
\bottomrule
\end{tabular}}
\end{table}

We observed that our NAS approach was able to find network candidates with slightly higher accuracy than all MCUNet variants tested, while offering an overall better memory footprint, especially in terms of ROM, at the cost of higher computational complexity. One reason for the higher computational complexity may be that MCUNet uses less computationally intensive depthwise-separable convolutions~\citep{chollet2017xception}, while our optimized architecture, i.e. ResNet18, uses regular convolutions. In addition, we noticed that the dynamic first-fit memory allocator of the DNN mapping tool we used often reported significantly lower RAM requirements for smaller input resolutions than MCUNet's patch-based strategy, while the opposite was true for larger input sizes.

We also report results for a version of MCUNet for CIFAR10 (mcunet-proxylessnas), which we explicitly searched using ProxylessNAS with the default configuration provided by the authors. We optimized for 300 steps using CIFAR10 as the dataset for both training and validation. Compared to our approach, which provides a Pareto front of trained and ready to use DNN models as a result, ProxylessNAS only outputs a single MCUNet configuration deemed optimal and estimates of the accuracy and FLOPS, which we then used for separate training. We observed that the actual performance of the configuration after training was relatively accurate, although slightly overestimated compared to the estimated values ($86.53\%$ and $27.16M$ estimated versus $82.33\%$ and $25.5M$ actual).

\subsubsection{Comparison with AMC and Automated Pruning}
\label{sec:automated_pruning}

\begin{figure}[t]
    \centering
    \includegraphics[width=.9\textwidth]{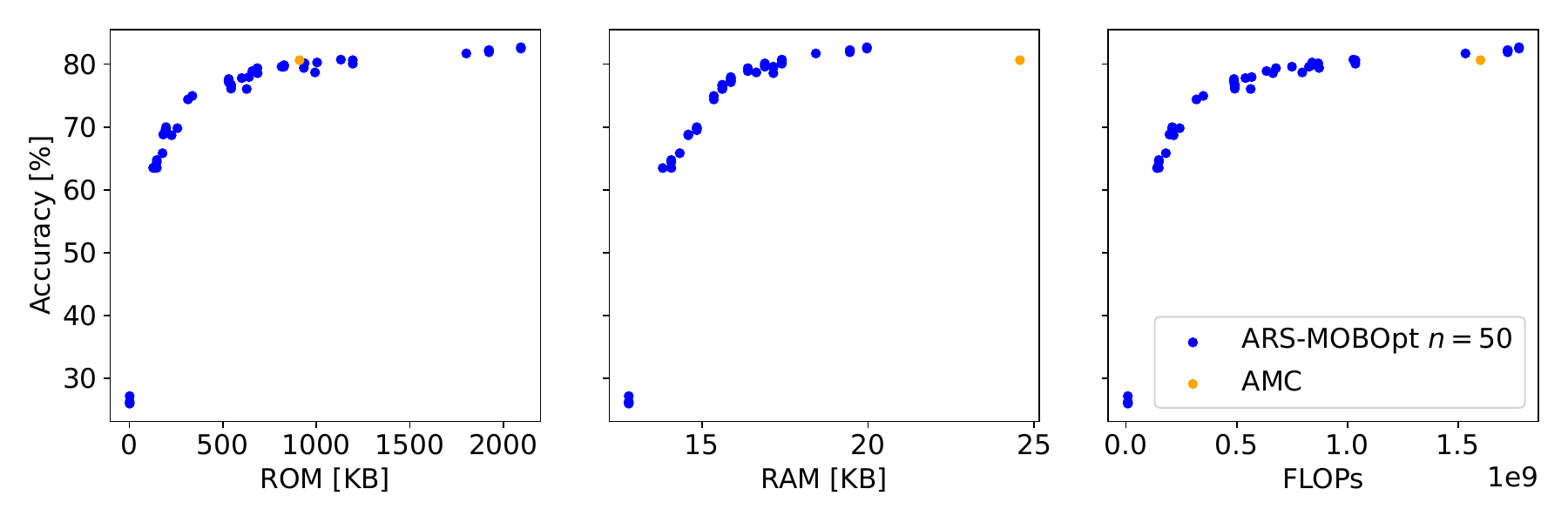}
    \caption{Optimization of MobileNet and CIFAR10 using ARS-MOBOpt (ours) and AMC. We ran our optimization for 50 samples while AMC by default only delivers a single tradeoff.}
    \Description{The feasible Pareto sets determined by ARS-MOBOpt for MobileNet and CIFAR10 compared to AMC. AMC produced only a single tradeoff that, while generally comparable in performance, is dominated by several of the feasible tradeoffs proposed by our approach in regards of accuracy, ROM, RAM, and FLOPs.}
    \label{fig:amc}
\end{figure}

Another approach to find an optimal pruning schedule for a given DNN architecture, is automated pruning~\citep{he2018amc,alwani2022decore,hirsch2022multi}. Most automated pruning algorithms are based on sensitivity analysis~\citep{han_deep_2016}. The idea of this method is that, given a well-trained baseline model, the sensitivity of each of its layers, i.e., the percentage of a layer's neurons that can be removed by pruning before the overall accuracy of the baseline DNN starts to degrade, can be easily queried iteratively layer by layer. Typically, during sensitivity analysis the baseline model is not retrained after pruning, making the method very time efficient. Once a sensitivity has been queried for each layer, a pruning schedule, i.e., a set of pruning rates for each layer, can be derived. This schedule can then be executed in a second training run, resulting in a well-trained and pruned DNN.

In the following, we compare our ARS-MOBOpt approach with AMC~\citep{he2018amc}, an automated pruning approach that uses a DDPG RL agent~\citep{lillicrap2015continuous} to iteratively query pruning rates for each layer given a well-trained baseline network. We search for pruned versions of MobileNetV1 on CIFAR10 using both our approach in blue and AMC in orange, see Fig.~\ref{fig:amc}. To make the results of our apporach and AMC directly comparable regarding memory and FLOPs, we use the same quantization and deployment method we described in Section~\ref{sec:algobench} for both the pruned model generated by AMC and ours. Since AMC proposes only a single pruning schedule, i.e., a single tradeoff, instead of a Pareto front of tradeoffs, only a single point is shown for AMC in Fig.~\ref{fig:amc}.

Considering only accuracy, we observed that AMC found a DNN pruning configuration that produced a DNN model comparable to the best model found by our approach. However, when also considering the other objectives, it can be seen that especially for RAM and FLOPs, our approach found significantly better tradeoffs than AMC. We believe that this is the result of two aspects: First, all approaches based on sensitivity analysis do not consider retraining after pruning. However, it is well known that a significant amount of accuracy initially lost by pruning can be recovered later by retraining~\citep{han_deep_2016, li_pruning_2017, deutel_deployment_2022}. Since sensitivity analysis does not consider retraining or pruning multiple layers together, it tends to produce conservative pruning schedules that overestimate the effect pruning has on the DNN. Second, AMC does not explicitly consider memory, only FLOPs in the reward of its DDPG agent, and also does not allow the explicit modeling of constraints. However, while this would explain why our framework found better tradeoffs between accuracy and memory, since, unlike AMC, it explicitly considers memory during optimization, it still does not fully explain why AMC did not find models with a similar number of FLOPs compared to the models proposed by our approach.

\section{Conclusion}
\label{section:conclusion}

We presented a NAS methodology for efficient multi-objective DNN optimization for microcontrollers based on Bayesian optimization, Augmented Random Search (ARS), and pruning. Since the objectives accuracy, ROM, RAM and FLOPs were expensive to evaluate and we were faced with a limited search budged, we focused on performing time-efficient optimization of both DNN hyper- and compression-parameters to enable an optimal deployment on microcontroller platforms. We provided results for two different problems considering two datasets and DNN architectures and showed empirically that our algorithm was able to yield a better feasible Pareto front compared to well-known Bayesian optimization strategies like ParEGO, TurBO or MorBO.


\bibliographystyle{ACM-Reference-Format}
\bibliography{references}

\appendix
\clearpage

\section{Complete Listing for UEA \& ICR Time Series Classification Benchmark}
\label{app:UEA-complete}

In Table~\ref{tab:uea_ucr} we provide all results we obtained for the multivariate datasets of the UEA \& UCR time series classification benchmark~\citep{bagnall16bakeoff}. We provide results for our ARS strategy (columns 3--7) and ParEGO (columns 8--12). For both strategies, we show the achieved hypervolume after 150 trials as well as the maximum accuracy. We also show the minimum memory and FLOPs required to achieve at least 70\% accuracy. If no trained networks were found that met these requirements (marked in gray), we show the memory and FLOPs requirements of the network that achieved the highest accuracy. We provide single objective reference results for InceptionTime \citep{fawaz2020inception} as taken from \cite{bagnall16bakeoff}.

\begin{table}[ht!]
    \centering
    \caption{Results for the multivariate datasets of the UEA \& UCR time series classification benchmark \citep{bagnall16bakeoff} for our ARS strategy (columns 3--7), ParEGO (columns 8--12) and InceptionTime (column 1) as a reference.}
    \label{tab:uea_ucr}
    \resizebox{\textwidth}{!}{
\begin{tabular}{l c c c c c c c c c c c}
\toprule
\textbf{Dataset} & \color{lightgray}{\textbf{InceptionTime}} & \multicolumn{5}{c}{\textbf{ARS}} & \multicolumn{5}{c}{\textbf{ParEGO}} \\
& \color{lightgray}{Acc. [\%]} & HV & Acc. [\%] & ROM [Kb] & RAM [Kb] & FLOPs & HV & Acc. [\%] & ROM [Kb] & RAM [Kb] & FLOPs \\
\midrule
Epilepsy & \color{lightgray}{98.55} & \textbf{0.97} & \textbf{97.81} & \num{120512} & \num{4920} & \num{3570578} & 0.93 & 94.16 & \textbf{\num[detect-weight]{66988}} & \textbf{\num[detect-weight]{3392}} & \textbf{\num[detect-weight]{2050650}} \\
BasicMotions & \color{lightgray}{100.00} & 0.96 & \textbf{97.50} & \num{131176} & \num{3000} & \num{2745720} & \textbf{0.97} & \textbf{97.50} & \textbf{\num[detect-weight]{86696}} & \textbf{\num[detect-weight]{2500}} & \textbf{\num[detect-weight]{1493195}} \\
Cricket & \color{lightgray}{98.61} & \textbf{0.96} & \textbf{97.22} & \num{100404} & \textbf{\num[detect-weight]{29924}} & \num{16020432} & 0.95 & 95.83 & \textbf{\num[detect-weight]{76068}} & \num{30520} & \textbf{\num[detect-weight]{10139994}} \\
ERing & \color{lightgray}{87.78} & \textbf{0.96} & \textbf{96.67} & \num{100256} & \num{1988} & \num{2025968} & 0.93 & 93.33 & \textbf{\num[detect-weight]{73196}} & \textbf{\num[detect-weight]{1228}} & \textbf{\num[detect-weight]{1002400}} \\
ArticularyWordRecognition & \color{lightgray}{98.33} & \textbf{0.94} & \textbf{95.64} & \num{150516} & \num{6180} & \num{4194267} & 0.92 & 92.36 & \textbf{\num[detect-weight]{80748}} & \textbf{\num[detect-weight]{5328}} & \textbf{\num[detect-weight]{2127501}} \\
UWaveGestureLibrary & \color{lightgray}{87.81} & \textbf{0.91} & \textbf{91.67} & \num{110256} & \num{6944} & \num{5779512} & \textbf{0.91} & \textbf{91.67} & \textbf{\num[detect-weight]{72052}} & \textbf{\num[detect-weight]{4092}} & \textbf{\num[detect-weight]{3027980}} \\
NATOPS & \color{lightgray}{96.11} & \textbf{0.86} & \textbf{87.22} & \num{129032} & \num{5784} & \num{2793175} & 0.78 & 78.33 & \textbf{\num[detect-weight]{100264}} & \textbf{\num[detect-weight]{4944}} & \textbf{\num[detect-weight]{1118016}} \\
SelfRegulationSCP1 & \color{lightgray}{83.96} & \textbf{0.79} & \textbf{79.85} & \num{104688} & \textbf{\num[detect-weight]{22400}} & \num{17814384} & 0.78 & 78.73 & \textbf{\num[detect-weight]{78052}} & \textbf{\num[detect-weight]{22400}} & \textbf{\num[detect-weight]{7147130}} \\
PEMS-SF & \color{lightgray}{--} & 0.76 & 81.50 & \textbf{\num[detect-weight]{137784}} & \textbf{\num[detect-weight]{554832}} & \textbf{\num[detect-weight]{7984536}} & \textbf{0.79} & \textbf{84.97} & \num{238016} & \num{555756} & \num{57178800} \\
Heartbeat & \color{lightgray}{58.05} & \textbf{0.73} & \textbf{74.51} & \num{81648} & \textbf{\num[detect-weight]{99224}} & \num{8139132} & 0.73 & 74.02 & \textbf{\num[detect-weight]{62532}} & \textbf{\num[detect-weight]{99224}} & \textbf{\num[detect-weight]{2793768}} \\
HandMovementDirection & \color{lightgray}{36.49} & \textbf{0.60} & 60.81 & \color{lightgray}{\num{171112}} & \color{lightgray}{\num{23564}} & \color{lightgray}{\num{10114533}} & 0.45 & 45.95 & \color{lightgray}{\num{99484}} & \color{lightgray}{\num{18588}} & \color{lightgray}{\num{9716270}} \\
FingerMovements & \color{lightgray}{56.00} & \textbf{0.59} & 60.00 & \color{lightgray}{\num{161940}} & \color{lightgray}{\num{6272}} & \color{lightgray}{\num{3100887}} & 0.58 & 59.00 & \color{lightgray}{\num{168232}} & \color{lightgray}{\num{6968}} & \color{lightgray}{\num{4340149}} \\
SelfRegulationSCP2 & \color{lightgray}{47.22} & \textbf{0.58} & 58.33 & \color{lightgray}{\num{105120}} & \color{lightgray}{\num{58708}} & \color{lightgray}{\num{39219912}} & 0.56 & 56.67 & \color{lightgray}{\num{183456}} & \color{lightgray}{\num{39156}} & \color{lightgray}{\num{24157260}} \\
Handwriting & \color{lightgray}{64.24} & 0.57 & 58.00 & \color{lightgray}{\num{208100}} & \color{lightgray}{\num{5424}} & \color{lightgray}{\num{4653128}} & \textbf{0.71} & 72.00 & \color{lightgray}{\num{130684}} & \color{lightgray}{\num{4752}} & \color{lightgray}{\num{5555856}} \\
EigenWorms & \color{lightgray}{--} & \textbf{0.55} & 60.16 & \color{lightgray}{\num{211472}} & \color{lightgray}{\num{449600}} & \color{lightgray}{\num{546991101}} & 0.50 & 53.91 & \color{lightgray}{\num{142288}} & \color{lightgray}{\num{683364}} & \color{lightgray}{\num{703130472}} \\
MotorImagery & \color{lightgray}{--} & 0.55 & 60.00 & \color{lightgray}{\num{148500}} & \color{lightgray}{\num{771000}} & \color{lightgray}{\num{83765403}} & \textbf{0.55} & 61.00 & \color{lightgray}{\num{158256}} & \color{lightgray}{\num{833956}} & \color{lightgray}{\num{190908344}} \\
EthanolConcentration & \color{lightgray}{23.19} & \textbf{0.53} & 54.02 & \color{lightgray}{\num{146820}} & \color{lightgray}{\num{59040}} & \color{lightgray}{\num{47050848}} & 0.49 & 49.43 & \color{lightgray}{\num{208636}} & \color{lightgray}{\num{45484}} & \color{lightgray}{\num{64709019}} \\
AtrialFibrillation & \color{lightgray}{20.00} & 0.53 & 53.33 & \color{lightgray}{\num{146720}} & \color{lightgray}{\num{21512}} & \color{lightgray}{\num{17560908}} & \textbf{0.59} & 60.00 & \color{lightgray}{\num{203488}} & \color{lightgray}{\num{16760}} & \color{lightgray}{\num{31350348}} \\
StandWalkJump & \color{lightgray}{40.00} & 0.49 & 50.00 & \color{lightgray}{\num{200744}} & \color{lightgray}{\num{93708}} & \color{lightgray}{\num{59756631}} & \textbf{0.66} & 66.67 & \color{lightgray}{\num{210240}} & \color{lightgray}{\num{86216}} & \color{lightgray}{\num{143958397}} \\
DuckDuckGeese & \color{lightgray}{--} & \textbf{0.39} & 46.00 & \color{lightgray}{\num{174460}} & \color{lightgray}{\num{1453136}} & \color{lightgray}{\num{34039926}} & 0.34 & 40.00 & \color{lightgray}{\num{144272}} & \color{lightgray}{\num{1452868}} & \color{lightgray}{\num{14406184}} \\
PhonemeSpectra & \color{lightgray}{--} & \textbf{0.21} & 21.42 & \color{lightgray}{\num{215612}} & \color{lightgray}{\num{15220}} & \color{lightgray}{\num{8697204}} & 0.19 & 19.49 & \color{lightgray}{\num{178176}} & \color{lightgray}{\num{12440}} & \color{lightgray}{\num{10222260}} \\
\bottomrule
\end{tabular}}
\end{table}

\section{Synthetic Minimization Problem}
\label{app:synth_example}

\begin{figure}[ht!]
    \centering
    \includegraphics[width=0.42\textwidth]{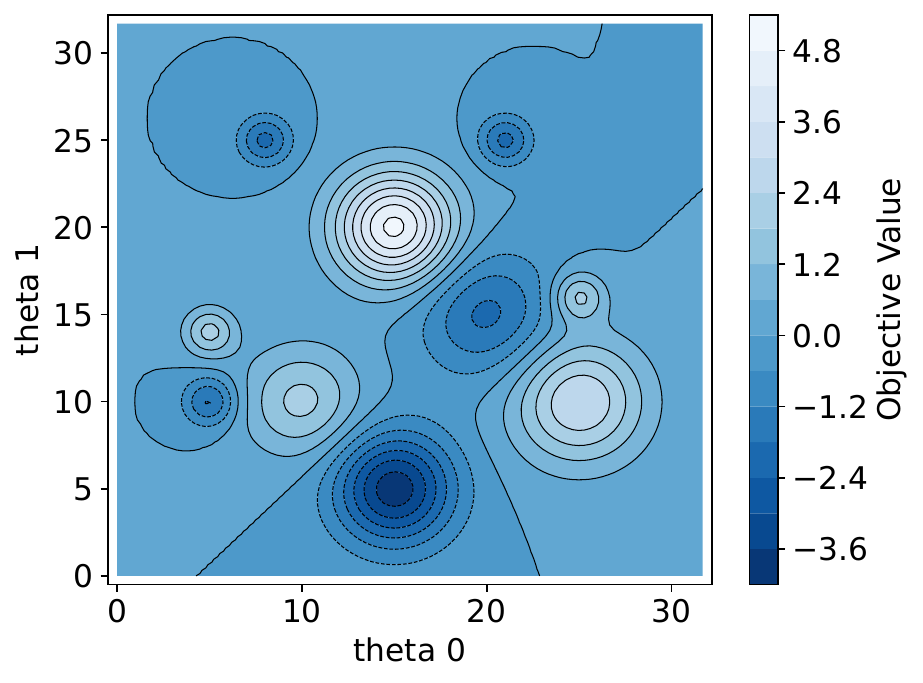}
    \caption{Topography of the synthetic minimization problem.}
    \Description{Topography of the synthetic minimization problem.}
    \label{fig:synth_gt}
\end{figure}

In Fig.~\ref{fig:synth_gt} we provide the topography of the synthteic minimization problem we consider in Section~\ref{sec:intuition}. The global minima can be found at $\theta_0 = 15$, $\theta_1 = 5$ and the problem was taken from~\cite{passino2005biomimicry}.

\section{PPO Hyperparameters}
\label{app:ppo_params}

Table~\ref{tab:ppo_params} shows the PPO hyperparameters found that maximize  the Hypervolume improvement of on of the multi-objective optimization problems discussed in Sec.~\ref{sec:ppo}, ResNet trained on CIFAR10, compared to the default hyperparameters\footnote{\url{https://stable-baselines3.readthedocs.io/en/master/modules/ppo.html}} of the algorithm.

\begin{table}[!ht]
    \caption{PPO hyperparameters found that maximize the Hypervolume improvement of ResNet trained on CIFAR10 (Optimized) compared to the default parameters (Default).}
    \label{tab:ppo_params}
    \begin{tabular}{lcccc}
        \toprule
        \textbf{Parameter} & \textbf{Search Space} & \textbf{Interval} & \textbf{Default} & \textbf{Optimized} \\
        \midrule
        batch\_size & categorical & $(32, 64, 128)$ & 64 & 64 \\
        clip\_range & categorical & $(0.1, 0.2, 0.3, 0.4)$ & 0.2 & 0.4 \\
        ent\_coef & uniform & $[1e^{-8}, 0.1]$ & 0.0 & 0.016 \\
        gae\_lambda & categorical & $(0.92, 0.95, 0.98)$ & 0.95 & 0.98 \\
        gamma & categorical & $(0.9, 0.95, 0.98, 0.99, 0.995)$ & 0.99 & 0.98 \\
        learning\_rate & uniform & $[1e^{-5}, 1]$ & $3e^{-4}$ & $5e^{-4}$ \\
        max\_grad\_norm & categorical & $(0.3, 0.5, 0.6, 0.7, 0.8, 0.9, 1, 2, 5)$ & 0.5 & 1.0 \\
        n\_epochs & categorical & $(5, 10, 20)$ & 10 & 20 \\
        n\_steps & categorical & $(16, 32, 64, 128)$ & 2048 & 32 \\
        vf\_coef & uniform & $[0, 1]$ & 0.5 & 0.24 \\
        \bottomrule
    \end{tabular}
\end{table}

\end{document}